\documentclass[journal,twoside,web]{ieeecolor}
\usepackage{tmi}
\usepackage{cite}
\usepackage{amsmath,amssymb,amsfonts}
\usepackage{algorithmic}
\usepackage{graphicx}
\usepackage{textcomp}
\usepackage{amssymb}
\usepackage{bbding}
\usepackage{tabularray}
\usepackage{colortbl}
\usepackage{booktabs}
\usepackage{multirow}
\definecolor{mycyan}{rgb}{0.925,0.941,0.961}
\usepackage[normalem]{ulem}
\usepackage{color}
\usepackage{fancyhdr}

\def\BibTeX{{\rm B\kern-.05em{\sc i\kern-.025em b}\kern-.08em
    T\kern-.1667em\lower.7ex\hbox{E}\kern-.125emX}}
\begin{document}
\title{Unified Multi-Modal Image Synthesis for Missing Modality Imputation}
\author{Yue Zhang, Chengtao Peng, Qiuli Wang, Dan Song, Kaiyan Li, S. Kevin Zhou,
\IEEEmembership{Fellow, IEEE}
% \author{Yue Zhang, \IEEEmembership{Fellow, IEEE}, Second B. Author, and Third C. Author, Jr., \IEEEmembership{Member, IEEE}
\thanks{\copyright~2024 IEEE. Personal use of this material is permitted. Permission from IEEE must be obtained for all other uses, in any current or future media, including reprinting/republishing this material for advertising or promotional purposes, creating new collective works, for resale or redistribution to servers or lists, or reuse of any copyrighted component of this work in other works.}
\thanks{This work was supported in part by Natural
Science Foundation of China under Grant 62271465, in part by Suzhou Basic Research Program under Grant SYG202338, in part by
Open Fund Project of Guangdong Academy of Medical Sciences, China (No. KY-KF202206), in part by the China Postdoctoral Science Foundation (2023M743410), and in part by the Jiangsu Funding Program for Excellent Postdoctoral Talent. \emph{(Corresponding author: S. Kevin Zhou.)}}
\thanks{Y. Zhang and K. Li are with the School of Biomedical Engineering, Division of Life Sciences and Medicine, University of Science and Technology of China (USTC), Hefei Anhui, 230026, China, and also with the Center for Medical Imaging, Robotics, Analytic Computing \& Learning (MIRACLE), Suzhou Institute for Advance Research, USTC, Suzhou Jiangsu, 215123, China (e-mail: yue\_zhang@ustc.edu.cn; kaiyan321@ustc.edu.cn).}
\thanks{S. K. Zhou is with the School of Biomedical Engineering, Division of Life Sciences and Medicine, University of Science and Technology of China (USTC), Hefei Anhui, 230026, China, the Center for Medical Imaging, Robotics, Analytic Computing \& Learning (MIRACLE), Suzhou Institute for Advance Research, USTC, Suzhou Jiangsu, 215123, China, the Key Laboratory of Precision and Intelligent Chemistry, USTC, Hefei Anhui, 230026, China, and also with the Key Laboratory of Intelligent Information Processing of Chinese Academy of Sciences (CAS), Institute of Computing Technology, CAS (e-mail: skevinzhou@ustc.edu.cn).}
\thanks{C. Peng is with University of Science and Technology of China, Hefei Anhui, 230026, China (e-mail: pct@mail.ustc.edu.cn).}
\thanks{Q. Wang is with the 7T Magnetic Resonance Translational Medicine Research Center, Department of Radiology, Southwest Hospital, Army Medical University (Third Military Medical University), Chongqing, 400038, China (e-mail: wangqiuli@tmmu.edu.cn).}
\thanks{D. Song is with the School of Electrical and Information Engineering, Tianjin University, Tianjin, 300072, China (e-mail: dan.song@tju.edu.cn).}
}

\maketitle

\begin{abstract}
Multi-modal medical images provide complementary soft-tissue characteristics that aid in the screening and diagnosis of diseases. However, limited scanning time, image corruption and various imaging protocols often result in incomplete multi-modal images, thus limiting the usage of multi-modal data for clinical purposes. To address this issue, in this paper, we propose a novel unified multi-modal image synthesis method for missing modality imputation. Our method overall takes a generative adversarial architecture, which aims to synthesize missing modalities from any combination of available ones with a single model. To this end, we specifically design a Commonality- and Discrepancy-Sensitive Encoder for the generator to exploit both modality-invariant and specific information contained in input modalities. The incorporation of both types of information facilitates the generation of images with consistent anatomy and realistic details of the desired distribution. Besides, we propose a Dynamic Feature Unification Module to integrate information from a varying number of available modalities, which enables the network to be robust to random missing modalities. The module performs both hard integration and soft integration, ensuring the effectiveness of feature combination while avoiding information loss. Verified on two public multi-modal magnetic resonance datasets, the proposed method is effective in handling various synthesis tasks and shows superior performance compared to previous methods.
\end{abstract}


\section{Introduction}
\label{sec:introduction}
\IEEEPARstart{M}{ulti-modal} medical images are widely adopted in disease screening and diagnosis due to their ability to provide complementary soft-tissue characteristics and diagnostic information. For instance, commonly acquired magnetic resonance (MR) sequences include T1-weighted, T2-weighted, post-contrast T1-weighted (T1Gd), and fluid-attenuated inversion recovery (FLAIR) images, each of which is considered as a distinct modality that highlights specific anatomy and pathology. Clinically, a combination of multiple modalities is often used to present pathological changes and assist clinicians in making accurate diagnoses. However, obtaining complete multi-modal images for each patient can be challenging due to factors such as limited scanning time, motion or artifact-induced image corruption, and the use of different imaging protocols~\cite{shen2020multi}. When dealing with incomplete data, it is undesirable to simply discard it as it often contains valuable information, and also infeasible to re-scan missing sequences for data completion due to the high cost of data acquisition. Therefore, multi-modal image synthesis (also known as data imputation) has been explored to generate missing modalities from limited available data, which has the potential to benefit downstream data analysis (i.e., segmentation~\cite{roy2010mr}, registration~\cite{iglesias2013synthesizing}), enhance the diagnostic accuracy of diseases (i.e., Alzheimer's disease~\cite{pan2018synthesizing}), and assist surgery planning~\cite{staartjes2021magnetic}.

Nowadays, data imputation approaches for medical images typically leverage generative models (\emph{e.g.,} generative adversarial networks~\cite{goodfellow2020generative}) to synthesize images through image translation, which can be grouped into \textbf{\emph{one-to-one synthesis}} methods
that generate one target contrast from a single modality
\cite{bowles2016pseudo,roy2016patch,roy2013magnetic,huang2016geometry,jog2015mr,ye2013modality,vemulapalli2015unsupervised,van2015cross,sevetlidis2016whole,li2014deep,nie2017medical,nie2018medical,wang2021realistic,dar2019image,yuan2020unified,luo2022adaptive,fei2022classification}, \textbf{\emph{many-to-one synthesis}} methods
that generate one target contrast from a certain set of modalities
\cite{jog2014random,jog2017random,lee2019collagan,li2019diamondgan,zhou2020hi,peng2021multi,olut2018generative,yang2019bi,hagiwara2019improving} and \textbf{\emph{unified synthesis}} methods
that generate any desired contrast from arbitrary combinations of available modalities
\cite{chartsias2017multimodal,sharma2019missing,shen2020multi,dalmaz2022resvit}. To cope with all possible synthesis scenarios, most one-to-one and many-to-one methods require training multiple generative models, 
which leads to multiplying training time. For instance, it needs $N\times(N-1)$ models for one-to-one methods to translate images between $N$ modalities. While a few one-to-one methods~\cite{yuan2020unified} can complete various one-to-one synthesis tasks within a single model, they could not be applicable to configurations where multiple input modalities exist. By contrast, unified synthesis provides a compact solution for missing data imputation, requiring only a single model to accommodate all input-output configurations.

Despite advancements achieved, current unified synthesis methods still have limitations in synthesizing tissues with fine structural details and sophisticated intensity variations (especially the tumor regions).
The underlying reasons lie in two aspects:
\textbf{(1)} Known methods typically rely on a single encoder~\cite{sharma2019missing} or a set of modality-specific encoders~\cite{chartsias2017multimodal} to process input modalities, which could not fully exploit the commonality and discrepancy information contained in multiple available modalities. This leads to limited information representation and incomplete synthesis of fine details and textures in synthetic images. \textbf{(2)} To ensure the network is robust to a varying number of available modalities, 
known unified methods derive unified latent features by a simple Max operation~\cite{chartsias2017multimodal} that only retains the maximum pixel value across modalities. However, this potentially leads to the loss of important information, such as subtle structural and intensity variations that are only present in a subset of modalities, and hence negatively impacts the synthesis performance.

To address the aforementioned issues, in this paper, we propose a novel unified multi-modal image synthesis network for missing contrast imputation. Our network employs a generative adversarial architecture that takes any combination of available modalities as input and generates synthetic images of missing modalities. To effectively extract commonality and discrepancy information contained in multiple available modalities, we develop a \textbf{\emph{Commonality- and Discrepancy-Sensitive Encoder}} (CDS-Encoder) for the generator. The CDS-Encoder comprises several modality-specific encoding streams that deal with the unique characteristics of each modality and a common encoding stream that copes with invariant structural features across different modalities. Such two types of information are progressively fused as the network goes deeper. Moreover, we propose a \textbf{\emph{Dynamic Feature Unification Module}} (DFUM) to derive unified latent features from multiple encoding streams, which is robust to any number of available modalities. The DFUM incorporates both hard and soft integration to ensure effective feature unification while minimizing information loss. Lastly, multiple decoding streams are used to decode the unified latent feature into different target modalities. During training, we employ a curriculum learning~\cite{bengio2009curriculum} strategy to ensure that the network learns efficiently from easier synthesis tasks and gradually adapts to harder tasks. Comprehensive experiments on two public multi-modal MR brain datasets demonstrate that our proposed method is effective in unified multi-modal image synthesis and outperforms known competing methods.

In summary, our contributions are four-fold:
\begin{itemize}
    \item We propose a novel unified multi-modal image synthesis network for missing modality imputation. Our network can take any combination of available modalities as input and generate the missing modalities, which is capable of handling all synthesis scenarios with a single model.
    \item We propose a CDS-Encoder for the generator, which can leverage both modality-invariant and modality-specific information from multiple input modalities. Modeling modality-invariant information aids in generating images with consistent anatomy and similar high-level features, while modality-specific information helps generate detailed soft-tissue characteristics.
    \item We propose a DFUM to adaptively integrate information from a varying number of available modalities. The DFUM incorporates both hard integration and soft integration to combine features in discrete and continuous manners
    while avoiding information loss.
    \item We evaluate our method on two public multi-modal MR brain datasets, including the BraTS dataset~\cite{bakas2017advancing,menze2014multimodal,bakas2018identifying} and IXI dataset. Experimental results demonstrate the effectiveness and superiority of our method compared to previous works.
\end{itemize}

\section{Related Works}
Medical image synthesis provides a promising solution to tackle the missing data problem, which has attracted increasing attention in recent years. In this section, we briefly review known synthesis methods by categorizing them into one-to-one, many-to-one, and unified synthesis methods.

\begin{figure*}[t]
\centerline{\includegraphics[width=\textwidth]{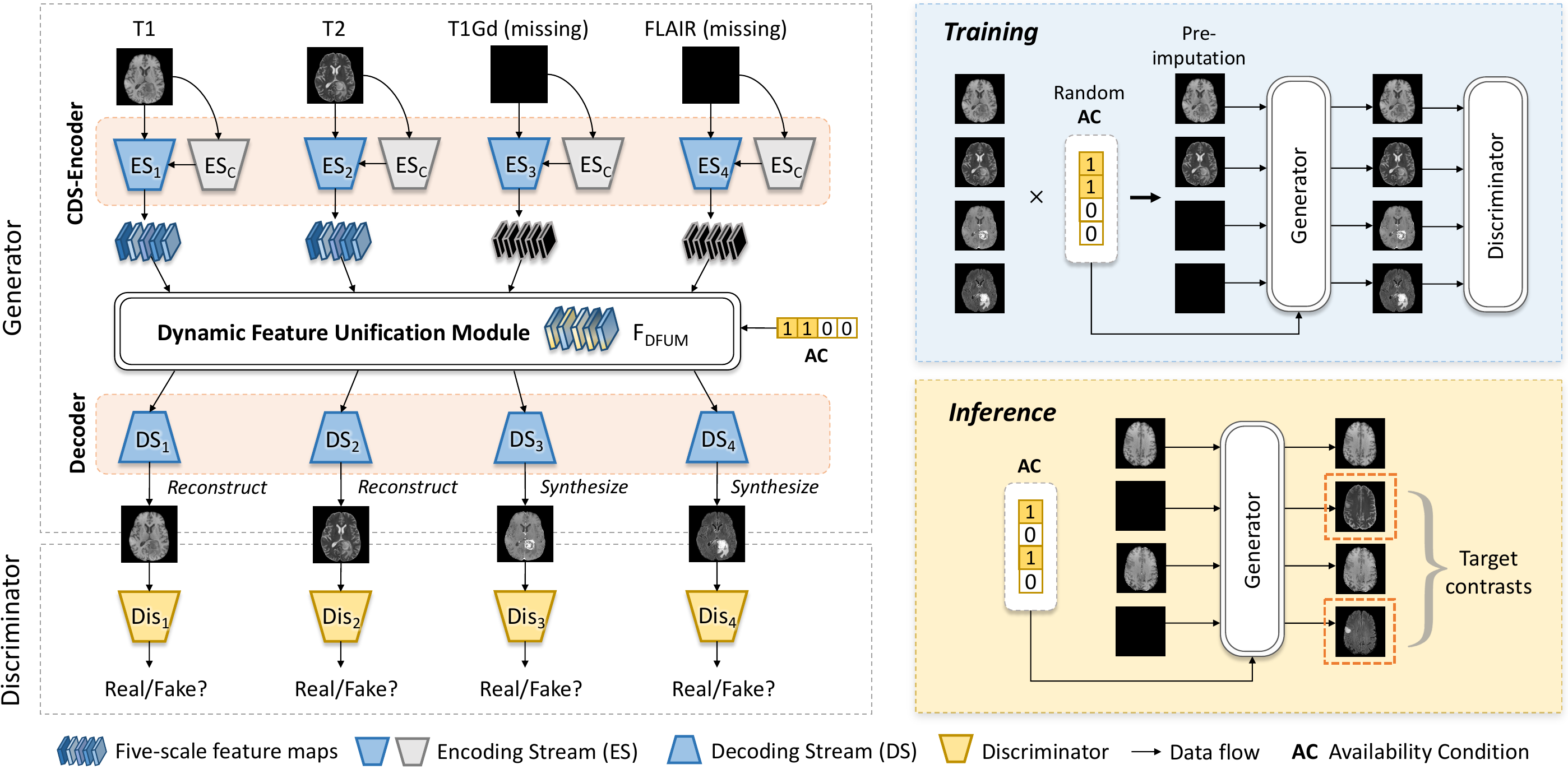}}
\caption{A schematic view of the proposed unified multi-modal image synthesis method.}
\label{fig_overview}
\end{figure*}

\textbf{One-to-One Synthesis:}
One-to-one synthesis methods take a single available contrast as input and generate a single target contrast. Earlier one-to-one synthesis studies were usually based on patch-based regression~\cite{bowles2016pseudo,roy2016patch}, sparse dictionary representation~\cite{roy2013magnetic,huang2016geometry} or atlas~\cite{jog2015mr,ye2013modality}, whose performance were subject to limited representation capability of hand-crafted features. With the development of deep learning techniques, convolutional neural networks (CNN) based methods have achieved immense success in one-to-one image synthesis. 
For instance, Sevetlidis \emph{et al.}~\cite{sevetlidis2016whole} proposed a deep encoder-decoder image synthesizer network for MR sequences.
Li \emph{et al.}~\cite{li2014deep} proposed a three-dimensional (3D) CNN to learn the mapping between volumetric positron emission tomography images and MR images. To further improve the performance, later studies used generative adversarial networks (GAN) for image synthesis, which could better learn high-frequency details by introducing an adversarial loss. For example, Nie \emph{et al.}~\cite{nie2017medical} proposed a context-aware GAN to model the nonlinear mapping from MR to CT. 
Dar \emph{et al.}\cite{dar2019image} proposed pGAN and cGAN for multi-contrast MR synthesis with a perceptual loss and a cycle-consistency loss. Yuan \emph{et al.}~\cite{yuan2020unified} proposed a GAN-based network that can learn various one-to-one mappings between multi-contrast MR images. Dalmaz et al.~\cite{dalmaz2024one} proposed to condition one-to-one mappings based on external guiding variables~\cite{liu2022undersampled} to unite different one-to-one synthesis tasks across sites. Tiwary et al.~\cite{tiwary2024cycle} employed twin energy-based models for unpaired image-to-image translation of medical images.
Chaudhary et al.~\cite{chaudhary2024lungvit} proposed a hierarchical transformer-based generator with squeeze-and-excitation decoder for inspiratory-to-expiratory intensity change in chest CT.
Most recently, diffusion models have emerged as a powerful tool for medical image analysis~\cite{shao2023diffuseexpand}, which is also developed for MR image synthesis. For instance, Ozbey et al.~\cite{ozbey2023unsupervised} proposed to use adversarial diffusion modeling for high-fidelity source-to-target MR synthesis. Wang et al.~\cite{wang2024mutual} proposed an unsupervised synthesis method based on diffusion models, which leverages the inherent statistical consistency of mutual information between different modalities.

\textbf{Many-to-One Synthesis:}
Many-to-one synthesis methods learn a mapping from multiple source contrasts to one target contrast. Like one-to-one synthesis methods, earlier many-to-one methods usually adopted patch-based regression, such as~\cite{jog2014random,jog2017random}. Recently, GAN-based methods have also been employed in many-to-one synthesis tasks.
For example, CollaGAN~\cite{lee2019collagan} and DiamondGAN~\cite{li2019diamondgan} incorporated a domain condition to guide the network to synthesize missing contrasts from available contrasts of MR brain images. Zhou \emph{et al.}~\cite{zhou2020hi} proposed a Hi-Net, which fused multi-scale information from different modalities to synthesize the target contrast. Yurt \emph{et al.}~\cite{yurt2021mustgan} proposed a MustGAN that shares a similar concept with our method in using common and distinct features. However, it requires step-wise training of individual networks to extract independent and shared features, and a joint network to combine these features. In contrast, the design of our network allows for end-to-end training, which can process different types of features simultaneously. Peng \emph{et al.}~\cite{peng2021multi} proposed a CACR-Net, which fused complementary information of multiple inputs from the output level and feature level to synthesize high-quality target-modality images.

\textbf{Unified Synthesis:}
Unified synthesis receives information from any combination of available source contrasts and generates the remnant missing contrasts in a single forward pass, which is a new research point in the field of medical image synthesis. Several studies~\cite{chartsias2017multimodal,sharma2019missing,shen2020multi,dalmaz2022resvit} have been proposed for this scenario. For example, Chartsias \emph{et al.}~\cite{chartsias2017multimodal} proposed a CNN-based multi-input multi-output model, which embedded all input contrasts into a shared latent space, and transformed the latent features into the target output modality with a decoder. Sharma \emph{et al.}~\cite{sharma2019missing} proposed an MM-GAN, which used information from all available contrasts to synthesize missing ones using a single encoding-decoding pathway. Shen \emph{et al.}~\cite{shen2020multi} proposed a GAN-based method with a disentanglement scheme to extract shared content encoding and separate style encoding across multiple domains. Dalmaz \emph{et al.}\cite{dalmaz2022resvit} used Transformer~\cite{han2022survey} to model long-range dependencies between different modalities for unified medical image synthesis. Liu et al.~\cite{liu2023one} and Hao et al.~\cite{hao2024qgformer} used transformers to encode hierarchical multi-contrast features, and then queried target contrasts for missing data imputation. As Chartsias's method~\cite{chartsias2017multimodal} has the most relevant design to our method, we emphasize the main similarities and differences. For similarities, both methods use independent encoders (decoders) to handle different modalities. The differences lie in 1) our method additionally designs a shared encoding stream to explicitly combine modality-invariant and -specific features; 2) our method employs a more flexible DFUM to integrate varying numbers of contrasts, while the other method uses a simple Max operator.

\section{Method}
\subsection{Unified Multi-Modal Synthesis Framework}
\label{sec_overview}
In this paper, we propose a novel unified synthesis framework for missing modality imputation from arbitrary combinations of available modalities. We mainly consider the unified synthesis task on four-modality MR sequences. However, it should be noted that our framework can be applied to multi-modal data with any number of modalities.

Fig.~\ref{fig_overview} gives a schematic view of the proposed method, which employs a generative adversarial architecture. 
Given randomly missing multi-modal inputs, and the Availability Conditions $AC=\{ac_i\in\{0,1\} \}_{i=1}^4$ of T1, T2, T1Gd, and FLAIR modality (with 0 denoting missing modalities, 1 denoting available modalities, $i$ denoting the $i^{th}$ modality), we pre-impute missing modalities with all-zeros~\cite{sharma2019missing} to ensure that the network always receives inputs with fixed four channels. \emph{\textbf{In the training stage}}, the generator employs a Commonality- and Discrepancy-Sensitive Encoder (CDS-Encoder) to model both modality-invariant and specific information and a Dynamic Feature
Unification Module (DFUM) to derive the unified latent features from input four channels, and finally decodes unified features into images of four modalities, including reconstructed images of available modalities and synthetic images of missing modalities. Subsequently, four discriminators distinguish between network-produced and real images. To ensure that the network is robust to various input configurations, the network is exposed to random missing scenarios in each training iteration with a randomly generated Availability Condition.
\emph{\textbf{In the inference stage}}, the generator takes the Availability Condition and zero-imputed four channels as input, and generates images of desired missing modalities. In the following subsections, we provide further details on our method.

\subsection{Generator}
The generator takes both available contrasts and zero-imputed missing contrasts as input and outputs images of four different modalities. Overall, the generator consists of an encoder, feature unification module, and decoder.
\subsubsection{Commonality- and Discrepancy-Sensitive Encoder}
Properly encoding the effective information contained in input modalities is crucial for synthesizing images with high fidelity. Particularly, \emph{common information} across all modalities guides the network to generate anatomically consistent target images, and \emph{distinct information} of each modality helps the network synthesize more accurate soft tissue details (especially for tumor region synthesis).
To fully exploit such two types of information, we propose a CDS-Encoder for the generator.

\begin{figure}[t]
\centerline{\includegraphics[width=0.47\textwidth]{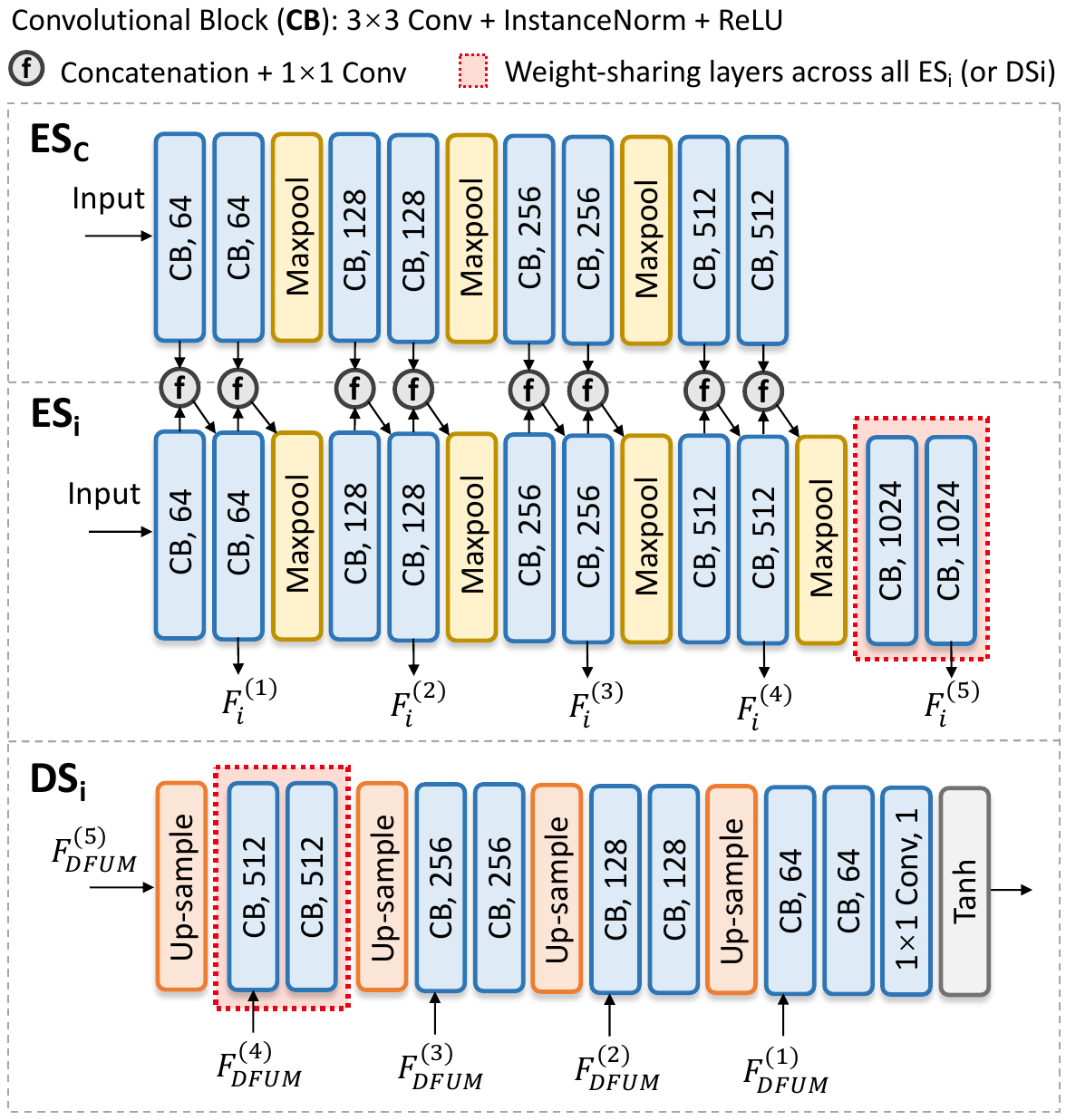}}
\caption{Illustration of the detailed structures of the common encoding stream ($ES_C$), modality-specific encoding stream ($ES_i$), and modality-specific decoding streams ($DS_i$).}
\label{fig_structure}
\end{figure}

\begin{figure}[t]
\centerline{\includegraphics[width=0.5\textwidth]{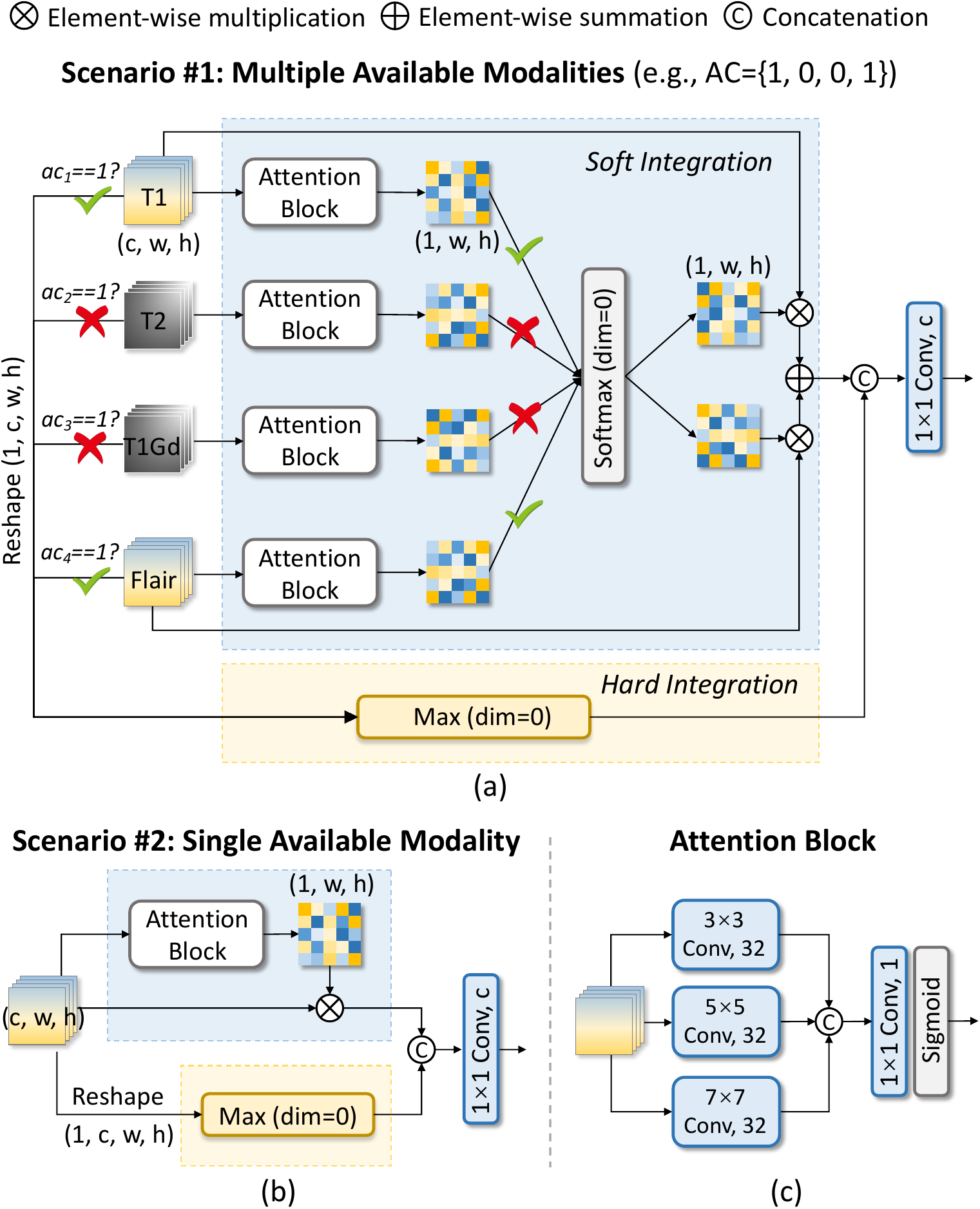}}
\caption{Illustration of the Dynamic Feature Unification Module (DFUM). (a) The scenario in which multiple modalities are available. (b) The scenario in which only a single modality is available. (c) The detailed structure of the attention block.}
\label{fig_afum}
\end{figure}

As depicted in Fig.~\ref{fig_overview}, CDS-Encoder incorporates four modality-specific encoding streams $\{ES_i\}_{i=1}^4$ (with the same structure and individual parameters) that cope with unique characteristics of each modality, along with a common encoding stream $ES_C$ that copes with shared and invariant features. As the network goes deeper, two types of features are gradually combined through fusion layers \textcircled{\footnotesize{f}} comprising a concatenation layer and a 1$\times$1 convolutional layer. Fig.~\ref{fig_structure} provides the detailed structures of $ES_i$ and $ES_C$. It should be noted that certain parameters are shared among four modality-specific encoding streams, as highlighted by the red dotted box in Fig.~\ref{fig_structure}. This design serves two purposes: firstly, it recognizes that features from different modalities should be similar at high levels; secondly, it reduces the memory usage of our network considering that we use multiple streams.

With the devised encoding architecture, the input image of $i^{th}$ modality is simultaneously fed into $ES_i$ and $ES_C$ in each forward pass. As such, the common encoding stream is shared among all input images so that it can concentrate on the common information across different modalities. Finally, the CDS-Encoder outputs multi-scale features of each modality $\{F_i^{(s)}\}_{i=1}^4,s\in\{1,2,3,4,5\}$ (\emph{i.e.,} four sets of five-scale features in total).

\subsubsection{Dynamic Feature Unification Module}
The key challenge in unified multi-modal synthesis is how to effectively integrate latent features from a varying number of available modalities. Existing unified synthesis methods typically use a Max function~\cite{chartsias2017multimodal} to obtain unified features that only preserve the maximum values across modalities, but this can result in a loss of valuable information (\emph{e.g.,} subtle structures and intensity variations present only in a subset of modalities) and negatively impact synthesis performance. To address this issue, we propose the DFUM that can dynamically integrate latent features from available modalities while retaining the unique characteristics of each modality. Fig.~\ref{fig_afum} illustrates the detailed structure of DFUM, which mainly consists of \emph{Hard Integration} and \emph{Soft Integration}. Given feature maps of the same scale from different modalities such as $\{F_i^{(1)}\} _{i=1}^4$ and Availability Condition $\{ac_i\}_{i=1}^4$, DFUM adaptively processes them according to various input configurations to make differential treatment and produces a unified feature $F_{DFUM}^{(1)}$. 

Specifically, Fig.~\ref{fig_afum} (a) shows \textbf{\emph{the scenario in which multiple modalities are available}}. In this scenario, Hard Integration preserves the maximum pixel value among different modalities through a Max operation, which retains the most informative features across modalities
and simply discards information with low response. This process can be formulated as:
\begin{equation}
    F_{hard}^{(s)}=Max(Concat\left \{ F_{i}^{(s)}\mid ac_{i}==1 \right \})
\end{equation}
where $i$ denotes the $i^{th}$ modality, $s$ denote the feature scale, and $Concat\{\cdot\}$ denotes the concatenation operation.
To enrich the representation and avoid information loss, Soft Integration employs attention mechanisms to learn the importance of each modality and weigh their pixel-wise contributions to the unified features. The used attention blocks are depicted in Fig.~\ref{fig_afum} (c), which employ convolutional layers with 3$\times$3, 5$\times$5, and 7$\times$7 kernels to compute spatial attention of each modality at different receptive fields. Noteworthy, missing modalities (with $ac_i=0$) are excluded from the integration process. The Soft Integration process can be formulated as:
\begin{equation}
    F_{soft}^{(s)}=\sum_{\substack{i=1\\ ac_{i}==1}} ^{4}W_{i}^{(s)}\otimes F_{i}^{(s)}
\end{equation}
\begin{equation}
    W^{(s)}=\sigma (Concat\left \{ Attn(F_{i}^{(s)})\mid ac_{i}==1 \right \})
\end{equation}
where $Attn(\cdot)$ denotes the attention block, and $\sigma(\cdot)$ denotes Softmax activation.
The final integrated feature is obtained by combining the Hard Integration and Soft Integration:
\begin{equation}
    F_{DFUM}^{(s)}=Conv_{1\times1}(Concat\left \{ F_{hard}^{(s)},F_{soft}^{(s)}\right \})
\end{equation}
where $Conv_{1\times1}$ denotes a convolutional layer with $1\times1$ kernel. To sum up, Hard Integration emphasizes the most important information from each modality while Soft Integration supplements discarded features after re-weighting, as such, DFUM can flexibly combine features from multiple available modalities in both discrete and continuous manners.

Fig.~\ref{fig_afum} (b) depicts \textbf{\emph{the scenario in which only a single modality is available}}. In this case, Soft Integration equals a self-attention module that encourages the focus on regions-of-interest (\emph{e.g.,} tumor regions), and Hard Integration equals an identity function:
\begin{equation}
    F_{hard}^{(s)}=F_{i}^{(s)}
\end{equation}
\begin{equation}
    F_{soft}^{(s)}=Attn(F_{i}^{(s)})
\end{equation}
in which the $i^{th}$ modality is the single available modality.
In other words, DFUM acts as a self-attention mechanism when processing a single available modality.

With this dynamic processing strategy, our DFUM facilitates both one-to-one (single modality available) and many-to-one (multiple modalities available) synthesis tasks. We plug five DFUMs in total into the generator and finally obtain five-scale unified features $\{F_{DFUM}^{(s)}\}_{s=1}^5$.

\subsubsection{Decoder}
The decoder utilizes four individual decoding streams $\{DS_i\}_{i=1}^4$ to decode five-scale unified features into images of four modalities. Each decoding stream is intended for synthesizing the contrast of a specific modality. For available modalities with $ac_i=1$, the decoder aims to reconstruct them while for missing modalities with $ac_i=0$, the decoder aims to synthesize soft-tissue contrasts with desired distributions. The detailed structure of $DS_i$ is illustrated in Fig.~\ref{fig_structure}, which progressively decodes and merges multi-scale features through sequential up-sampling and convolutional layers. Similar to the encoder, certain parameters are shared among all decoding streams (see the red dotted box in Fig.~\ref{fig_structure}).

\subsubsection{Loss Function}
We employ a hybrid loss to optimize the generator, consisting of a synthesis loss, a reconstruction loss, and an adversarial loss. 
Concretely, the synthesis loss $\mathcal{L}_{syn}$ is a pixel-wise L1 loss that measures the difference between synthetic images (of missing modalities) and real images:
\begin{equation}
  \mathcal{L}_{syn}=\sum_{i=1}^{4}(1-ac_i)\cdot L_1(\hat{Y}_i, Y_i),
\end{equation}
where $\hat{Y}_i$ denotes the network-produced image of the $i^{th}$ modality, $Y_i$ denotes the ground-truth image. $ac_i$ is the availability of the $i^{th}$ modality, which is randomly generated and may be different in each iteration. $(1-ac_i)$ ensures that only missing modalities are involved.

The reconstruction loss $\mathcal{L}_{rec}$ is a pixel-wise L1 loss measuring the difference between reconstructed images (of available modalities) and real images:
\begin{equation}
  \mathcal{L}_{rec}=\sum_{i=1}^{4}ac_i \cdot L_1(\hat{Y}_i, Y_i),
\end{equation}
where $ac_i$ ensures that only available modalities are involved.

The adversarial loss $\mathcal{L}_{adv}$ is a least square loss (L2 loss)~\cite{mao2017least} that fools the discriminators. Upon convergence, the generator produces realistic images that are indistinguishable from real images, enabling the network to better model high-frequency information and produce images that are coincident
with the target distribution:
\begin{equation}
    \mathcal{L}_{adv}=\sum_{i=1}^{4} (1-ac_i) \cdot (L_2(Dis_i(\hat{Y}_i)-1)+L_2(Dis_i(Y_i))),
\end{equation}
where $Dis_i$ denotes the discriminator of the $i^{th}$ modality.

So far, the overall loss of the generator is formulated as:
\begin{equation}
    \mathcal{L}_{Gen}=\lambda_1\mathcal{L}_{syn}+\lambda_2\mathcal{L}_{rec}+\lambda_3\mathcal{L}_{adv}.
\end{equation}
where $\lambda_1$, $\lambda_2$, and $\lambda_3$ are trade-off parameters for each loss item, which are selected through grid search.

\subsection{Discriminator}
Our network employs four discriminators $\{Dis_i\}_{i=1}^4$ to distinguish between real and synthetic images of each modality. The network structure of each discriminator follows PatchGAN~\cite{isola2017image}. Distinct from vanilla GANs that distinguish real and fake images from the whole-image level, PatchGAN conducts patch-level discrimination for input images. This can facilitate modeling local features and synthesizing realistic local details.

We use a least square loss to optimize the discriminators: 
\begin{equation}
    \mathcal{L}_{Dis}=\sum_{i=1}^{4} (1-ac_i) \cdot (L_2(Dis_i(\hat{Y}_i))+L_2(Dis_i(Y_i)-1)).
\end{equation}

During training, discriminators make efforts to tell apart synthetic images from real ones, which engage in adversarial training with the generator.

\subsection{Training Scheme}
\label{training_scheme}
To make the network robust to any missing data scenarios when testing, we expose the network to various input configurations during training. This is achieved by masking out a subset of modalities using a random Availability Condition in each iteration as stated in Section~\ref{sec_overview}.
However, since different missing data imputation tasks pose varying levels of difficulty, we introduce a Curriculum Learning~\cite{bengio2009curriculum} based training scheme. Following MM-GAN~\cite{sharma2019missing}, we show the network easy samples (missing one contrast) for the first 10 epochs, followed by moderate samples (missing two contrasts) for the next 10 epochs, and finally hard samples (missing three contrasts) for another 10 epochs. After that, we randomly set the input scenarios for the remaining epochs. This training mechanism enables the network to efficiently learn from easier tasks and gradually adapt to harder tasks~\cite{sharma2019missing}, which helps to improve the convergence speed and generalization ability.

Our network is implemented using Pytorch and deployed on an NVIDIA A100 GPU. We use stochastic gradient descent as the training optimizer with $momentum=0.9$, $weight\_decay=0.0001$. The learning rate is initially set to $2\times10^{-4}$ for the first 50 epochs and linearly decays to zero afterward. The total number of training epochs is 200 and the batch size is 32. 
For hyper-parameters in the loss functions, we considered $\lambda_1$ in $\left \{50, 100, 150, 200\right \}$, $\lambda_2$ in $\left \{0,10,20,...,60\right \}$ and $\lambda_3$ = 1. The optimal parameters $\lambda_1$=100, $\lambda_2$=30, $\lambda_3$=1, are selected through grid search.

\section{Materials and Experiments}
\label{sec_results}
\begin{figure*}[t]
\centerline{\includegraphics[width=\textwidth]{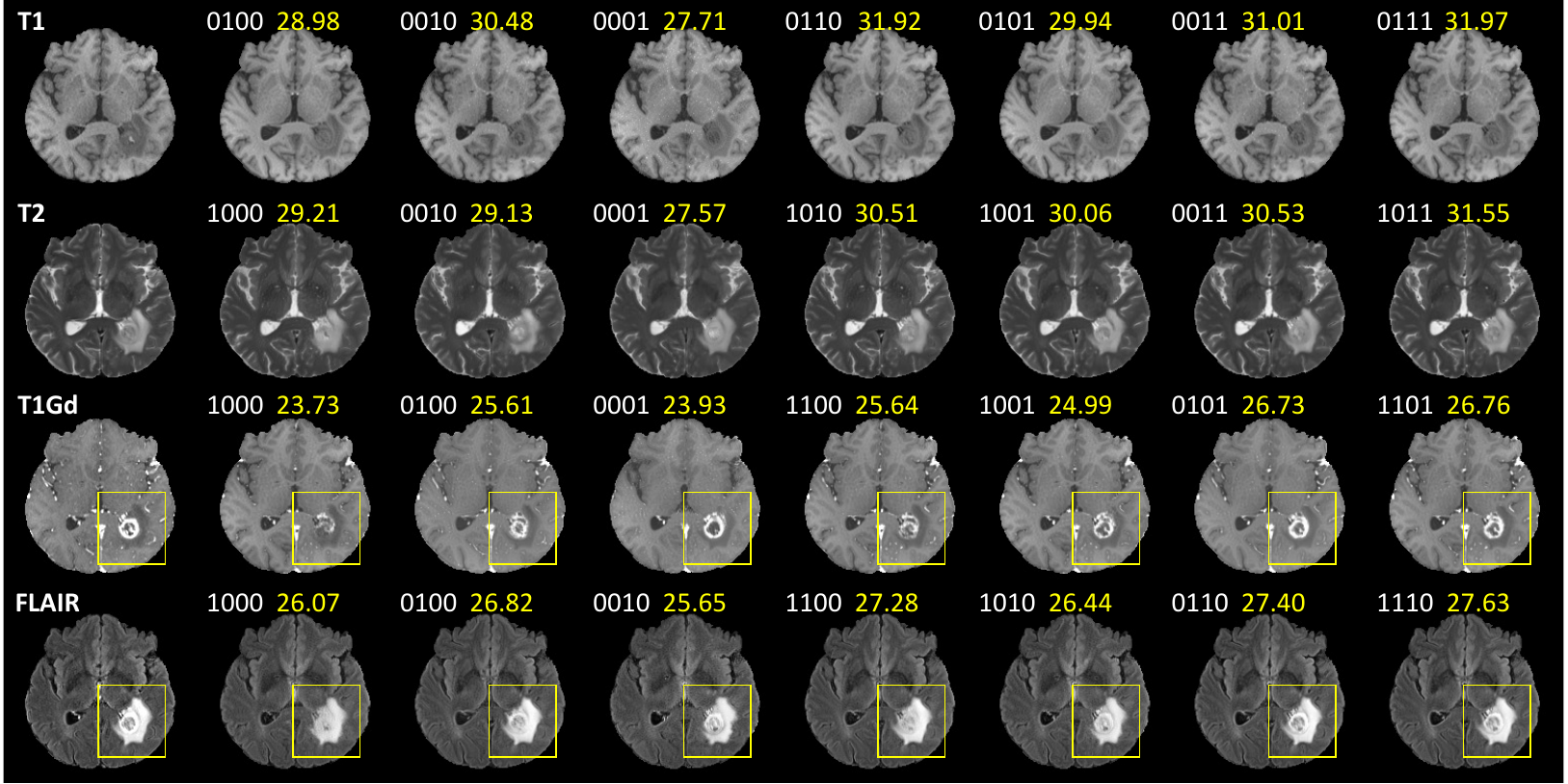}}
\caption{Visual examples of synthetic images produced by our method on the BraTS dataset. The four-bit digits represent the Availability Conditions of T1, T2, T1Gd, and FLAIR modalities, in which ``0" represents the ``missing" modality and ``1" represents the ``available" modality. Yellow boxes emphasize the obvious difference between images. The yellow decimals represent PSNR values.}
\label{fig_res_brats}
\end{figure*}

\subsection{Materials}
\label{sec_material}
To validate the effectiveness of our method, we conduct comprehensive experiments on two public multi-modal MR datasets including the BraTS 2019 dataset~\cite{bakas2017advancing,menze2014multimodal,bakas2018identifying} and the IXI dataset \footnote{https://brain-development.org/ixi-dataset}.

\subsubsection{BraTS Dataset}
The BraTS 2019 dataset comprises MR scans of glioblastoma and lower-grade glioma collected from 19 institutions. Each study includes skull-stripped and co-registered MR images of T1-weighted, T2-weighted, T1Gd, and FLAIR modalities. In our experiments, we randomly select 250, 20, and 20 artifact-free subjects for training, evaluation, and testing, respectively. We only use the middle 80 axial slices from each subject, which are cropped to a size of 192$\times$192 from the center region.

\subsubsection{IXI Dataset}
The IXI dataset comprises MR scans of healthy volunteers collected from three hospitals in London. Each study includes non-skull-stripped MR images of T1-, T2-, and PD-weighted modalities. We randomly select 170, 15, and 15 subjects for training, evaluation and testing. To prepare the images for analysis, we register the T1 and T2 volumes to the PD volumes using affine transformation (implemented through ANTsPy\footnote{https://github.com/ANTsX/ANTsPy}). For each subject, we select 60 artifact-free brain tissue slices from the middle of the image stack, which are then cropped to a size of 224$\times$224 from the center region.

\begin{table*}
\renewcommand\arraystretch{1.0}
\centering
\caption{Quantitative results of our method on the BraTS dataset.}
\label{tb_res_brats}
\begin{tabular}{cccccccc} 
\hline \hline
\multicolumn{4}{c}{Available modalities} & \multicolumn{4}{c}{Results [mean \textbf{PSNR} (std), mean \textbf{SSIM} (std)]}                                                                \\ 
\hline
T1 & T2 & T1Gd & FLAIR                & T1                          & T2                          & T1Gd                        & FLAIR                        \\ 
\hline
  &   &     & \checkmark                    & 27.38 (1.29), 0.938 (0.013) & 26.52 (1.22), 0.924 (0.021) & 24.65 (0.99), 0.883 (0.024) & -                            \\
  &   & \checkmark    &                     & 31.09 (1.84), 0.969 (0.020)  & 27.66 (1.08), 0.941 (0.017) & -                           & 25.52 (1.12), 0.882 (0.024)  \\
  & \checkmark  &     &                     & 28.88 (1.38), 0.958 (0.018) & -                           & 25.96 (0.94), 0.915 (0.024) & 26.81 (1.56), 0.903 (0.028)  \\
\checkmark  &   &     &                     & -                           & 27.78 (1.16), 0.944 (0.017) & 26.57 (1.47), 0.921 (0.030)  & 25.94 (1.08), 0.891 (0.023)  \\
  &   & \checkmark    & \checkmark                    & 31.47 (1.80), 0.972 (0.017)  & 28.78 (1.27), 0.954 (0.014) & -                           & -                            \\
  & \checkmark  &     & \checkmark                    & 29.47 (1.28), 0.963 (0.013) & -                           & 26.55 (1.11), 0.923 (0.023) & -                            \\
\checkmark  &   &     & \checkmark                    & -                           & 28.68 (1.29), 0.954 (0.014) & 27.08 (1.41), 0.928 (0.028) & -                            \\
  & \checkmark  & \checkmark    &                     & 31.86 (2.03), 0.974 (0.019) & -                           & -                           & 27.16 (1.60), 0.911 (0.026)   \\
\checkmark  &   & \checkmark    &                     & -                           & 28.48 (1.14), 0.952 (0.015) & -                           & 26.22 (1.21), 0.898 (0.023)  \\
\checkmark  & \checkmark  &     &                     & -                           & -                           & 27.68 (1.49), 0.938 (0.025) & 27.32 (1.57), 0.913 (0.026)  \\
  & \checkmark  & \checkmark    & \checkmark                    & \textbf{31.94} (1.94), \textbf{0.975} (0.017) & -                           & -                           & -                            \\
\checkmark  &   & \checkmark    & \checkmark                    & -                           & \textbf{29.28} (1.28), \textbf{0.959} (0.013) & -                           & -                            \\
\checkmark  & \checkmark  &     & \checkmark                    & -                           & -                           & \textbf{27.93} (1.51), \textbf{0.940} (0.025)  & -                            \\
\checkmark  & \checkmark  & \checkmark    &                     & -                           & -                           & -                           & \textbf{27.39} (1.59), \textbf{0.915} (0.026)  \\
\hline \hline
\end{tabular}
\end{table*}

In this study, both two datasets are normalized using mean normalization to ensure comparable ranges of voxel intensities across subjects. 

\begin{figure}[t]
\centerline{\includegraphics[width=0.5\textwidth]{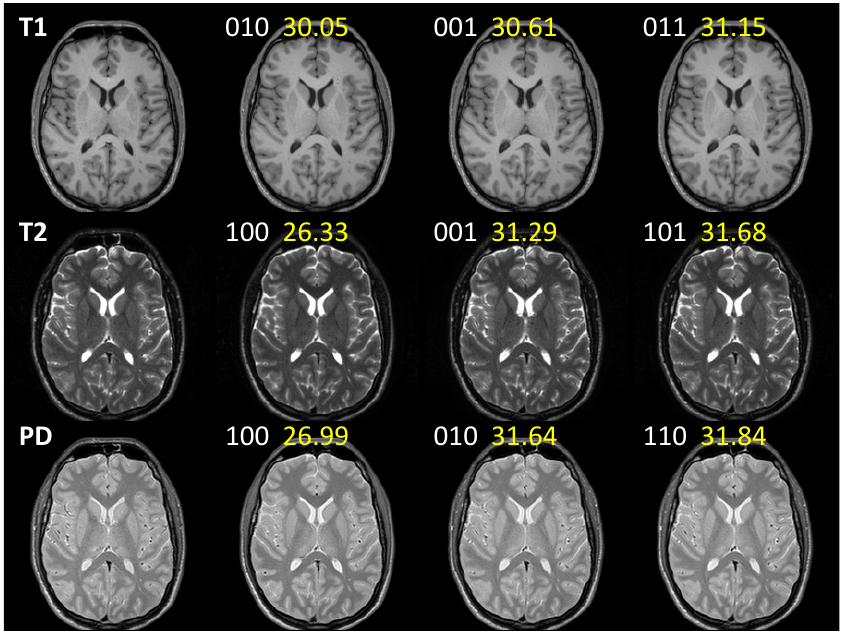}}
\caption{Visual examples of synthetic images produced by our method on the IXI dataset. The three-bit digits represent the Availability Conditions of T1, T2, and PD modalities, in which ``0" represents ``missing" modality and ``1" represents ``available" modality. The yellow decimals represent PSNR values.}
\label{fig_res_ixi}
\end{figure}

\subsection{Competing Methods and Evaluation Metrics}
To demonstrate the superiority of our method, we compare it with state-of-the-art methods in various synthesis tasks. Specifically, in the \emph{one-to-one synthesis} scenario, two competing methods are considered: \textbf{pGAN}~\cite{dar2019image} and \textbf{Pix2Pix}~\cite{isola2017image}; In the \emph{many-to-one synthesis} scenario, \textbf{Hi-Net}~\cite{zhou2020hi} is considered as the competing method; Finally, in the \emph{unified synthesis} scenario, we consider three competing methods: \textbf{MM-Synthesis}~\cite{chartsias2017multimodal}, \textbf{MM-GAN}~\cite{sharma2019missing} and \textbf{MM-Transformer}~\cite{liu2023one}. Note that MM-GAN and MM-Transformer are reproduced strictly according to their papers while other comparison methods are implemented using open-source codes. For the competing methods, we use hyper-parameters of losses provided in their papers. All the methods are trained by 200 epochs, with an initial learning rate of $2\times10^{-4}$ that linearly decays to 0 after 50 epochs, and a batch size of 32.

To quantitatively assess the performance of synthesis methods, we adopt two commonly-used evaluation metrics: peak signal-to-noise ratio (PSNR) and structural similarity index (SSIM).
Given a synthetic volume $\hat{y}$ and the ground-truth volume $y$, PSNR is defined as:
\begin{equation}
    PSNR(\hat{y},y)=10log_{10}{\frac{max^2(\hat{y},y)}{|\Omega|^{-1}\sum \left \| \hat{y}-y \right \|_2^2}}
\end{equation}
where $|\Omega|$ is the voxel numbers of $y$, $max^2(\hat{y},y)$ is the maximal intensity value of $\hat{y}$ and $y$.
SSIM is defined as:
\begin{equation}
SSIM(\hat{y},y)=\frac{(2\mu_{\hat{y}}\mu_{y}+c_1 )(2\sigma _{\hat{y}y}+c_2)}{(\mu_{\hat{y}}^2+\mu_{y}^2+c_1)(\sigma_{\hat{y}}^2+\sigma_{y}^2+c_2)}
\end{equation}
where $\mu_{y}$, $\sigma_{y}$ is the mean and variance of $y$, $\sigma _{\hat{y}y}$ is the covariance between $\hat{y}$ and $y$.
The higher value of PSNR and SSIM indicates better image quality.
Besides, we calculate the $p$-value using non-parametric Wilcoxon signed-rank test to assess the statistical significance of our method in comparison to other methods.

\begin{table}
\renewcommand\arraystretch{1.0}
\centering
\caption{Quantitative results of our method on the IXI dataset.}
\label{tb_res_ixi}
\begin{tabular}{cccccc} 
\hline \hline
\multicolumn{3}{c}{Available modalities} & \multicolumn{3}{c}{Results [mean \textbf{PSNR} (std), mean \textbf{SSIM} (std)]}                                                                                                                                                                \\ 
\hline
T1 & T2 & PD                          & T1                                                                    & T2                                                                    & PD                                                                     \\ 
\hline
% \rowcolor{mycyan}
  &   & \checkmark                           & \begin{tabular}[c]{@{}l@{}}29.31 (0.53), \\0.969 (0.005)\end{tabular} & \begin{tabular}[c]{@{}l@{}}30.93 (0.48), \\0.943 (0.027)\end{tabular} & -                                                                      \\ 
% \hline
% \cline{4-6}
  & \checkmark  &                            & \begin{tabular}[c]{@{}l@{}}28.48 (0.51), \\0.964 (0.007)\end{tabular} & -                                                                     & \begin{tabular}[c]{@{}l@{}}31.07 (0.46), \\0.977 (0.004)\end{tabular}  \\ 
% \hline
% \cline{4-6}
% \rowcolor{mycyan}
\checkmark  &   &                            & -                                                                     & \begin{tabular}[c]{@{}l@{}}26.43 (0.39), \\0.911 (0.026)\end{tabular} & \begin{tabular}[c]{@{}l@{}}26.96 (0.48), \\0.950 (0.008)\end{tabular}   \\ 
% \hline
% \cline{4-6}
  & \checkmark  & \checkmark                           & \begin{tabular}[c]{@{}l@{}}\textbf{29.50} (0.51), \\\textbf{0.971} (0.006)\end{tabular} & -                                                                     & -                                                                      \\ 
% \hline
% \cline{4-6}
% \rowcolor{mycyan}
\checkmark  &   & \checkmark                           & -                                                                     & \begin{tabular}[c]{@{}l@{}}\textbf{31.32} (0.49), \\\textbf{0.948} (0.025)\end{tabular} & -                                                                      \\ 
% \hline
% \cline{4-6}
\checkmark  & \checkmark  &                            & -                                                                     & -                                                                     & \begin{tabular}[c]{@{}l@{}}\textbf{31.86} (0.41), \\\textbf{0.980} (0.004)\end{tabular}   \\
\hline \hline
\end{tabular}
\end{table}

\subsection{Results of the Proposed Method}
In Fig.~\ref{fig_res_brats} and Fig.~\ref{fig_res_ixi}, we present examples of synthetic images produced by our method on the BraTS and IXI datasets. The four-bit (or three-bit) digits in the figures indicate the Availability Conditions of T1, T2, T1Gd, and FLAIR modalities (or T1, T2, and PD modalities), with ``1" representing available modality and ``0" representing missing modality. Our results demonstrate that synthesizing a target contrast using more available contrasts produces images with better quality. The combination of complementary information from multiple modalities is particularly crucial for synthesizing tumor regions with accurate shapes and realistic textures. For example, when synthesizing T1Gd images with glioblastoma (see the third row in Fig.~\ref{fig_res_brats}), using only T1 information leads to unsatisfactory results with incomplete and improperly enhanced tumor regions. However, extra combining T2 or FLAIR information greatly improves the synthesis quality as these two modalities provide clear demarcation of tumors. Table~\ref{tb_res_brats} and Table~\ref{tb_res_ixi} present the quantitative results of our method with different input-output configurations on the two datasets. The tables show that synthesizing the target modality using the largest number of available modalities achieves the best performance as measured by PSNR and SSIM, which is consistent with the qualitative results.

\begin{figure}[t]
\centerline{\includegraphics[width=0.5\textwidth]{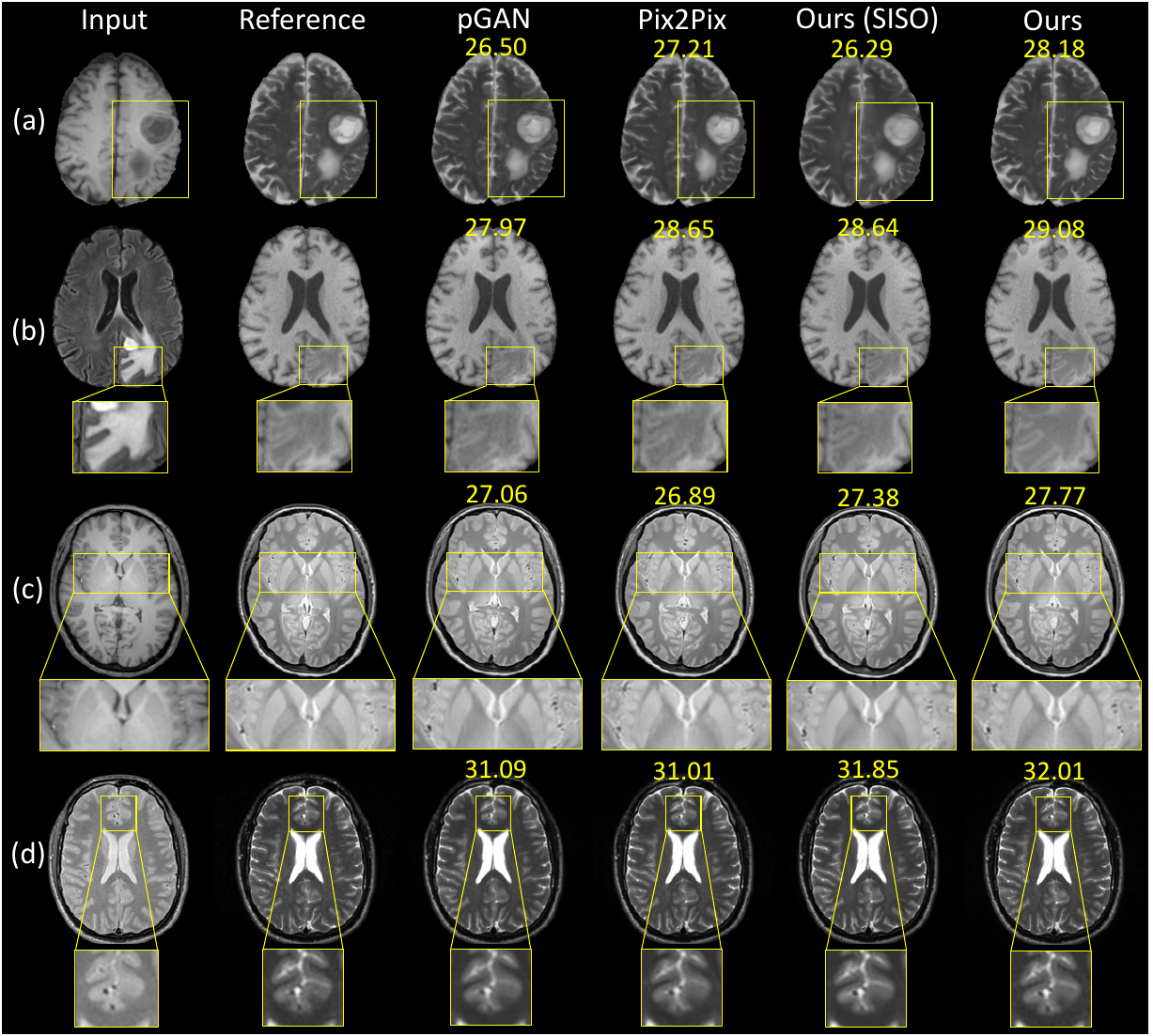}}
\caption{Visual examples of comparison results in one-to-one synthesis tasks on two datasets: (a) T1$\rightarrow$T2 and (b) FLAIR$\rightarrow$T1 in the BraTS dataset; (c) T1$\rightarrow$PD and (d) PD$\rightarrow$T2 in the IXI dataset. Yellow boxes emphasize distinctions between images, in which regions with subtle differences are enlarged for a better view. The yellow decimals represent PSNR values.}
\label{fig_res_one2one}
\end{figure}

\subsection{Comparison with State-of-the-Art Methods}
Our proposed unified synthesis method is capable of handling synthesis tasks with various input-output configurations. In this section, we specifically compare our method with state-of-the-art methods in one-to-one, many-to-one, and unified synthesis tasks, respectively.

\begin{figure*}[t]
\centerline{\includegraphics[width=\textwidth]{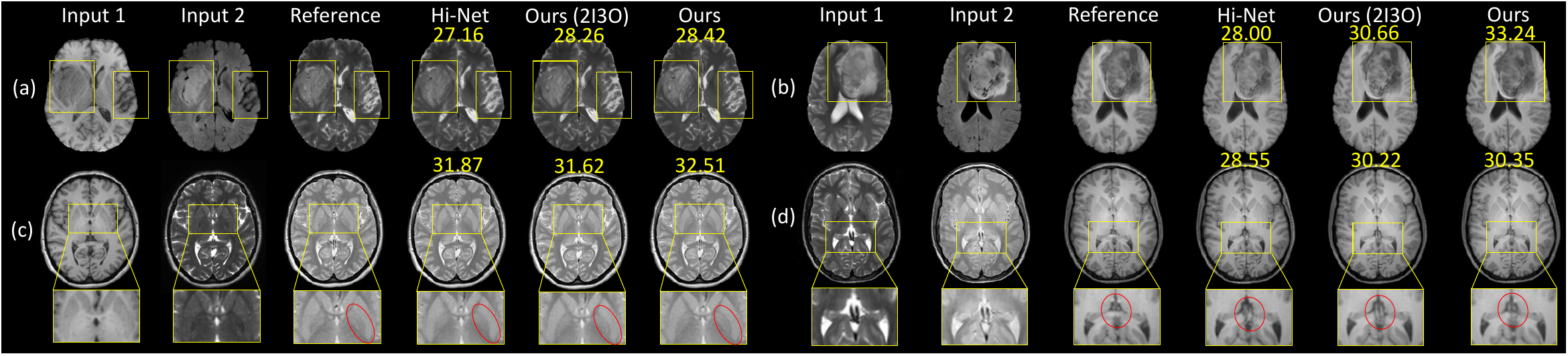}}
\caption{Visual examples of comparison results in many-to-one synthesis tasks on two datasets: (a) T1+FLAIR$\rightarrow$T2 and (b) T2+FLAIR$\rightarrow$T1 in the BraTS dataset; (c) T1+T2$\rightarrow$PD and (d) T2+PD$\rightarrow$T1 in the IXI dataset. Yellow boxes emphasize distinctions between images, in which regions with subtle differences are enlarged for a better view, the red ellipse emphasizes the streaking noises, and the yellow decimals represent PSNR values.}
\label{fig_res_many2one}
\end{figure*}

\begin{figure*}[t]
\centerline{\includegraphics[width=0.9\textwidth]{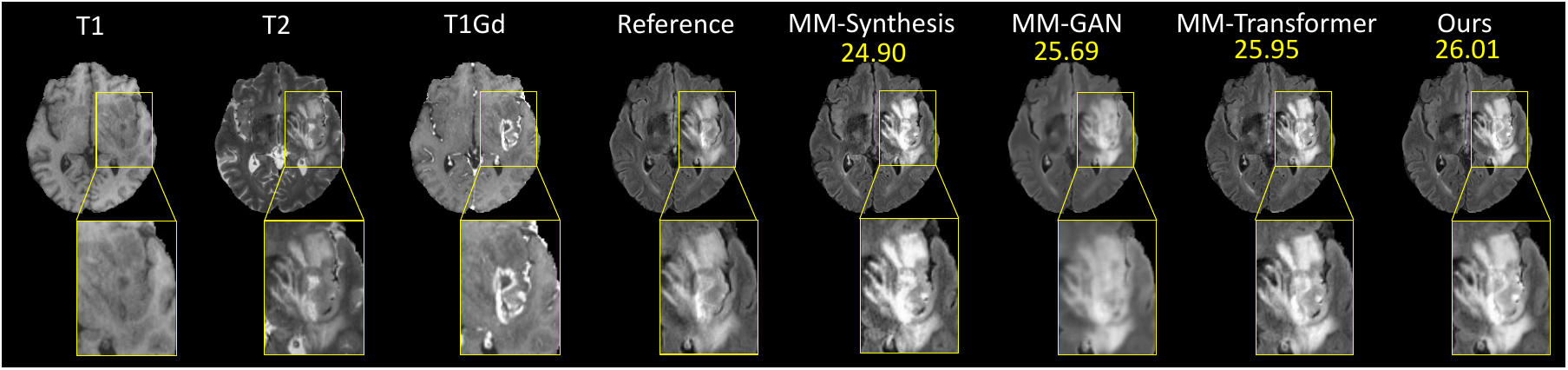}}
\caption{Representative examples (T1+T2+T1Gd$\rightarrow$FLAIR) of unified synthesis on the BraTS dataset. Yellow boxes emphasize distinctions between images, which are enlarged for a better view, and the yellow decimals represent PSNR values.}
\label{fig_res_uni_brats}
%\vspace{-1.5 mm}
\end{figure*}

\begin{table}
\renewcommand\tabcolsep{2.8pt}
\renewcommand\arraystretch{1.0}
\centering
\caption{Quantitative comparison with one-to-one synthesis methods in various tasks on the BraTS dataset (T1$\rightarrow$T2 and FLAIR$\rightarrow$T1) and IXI dataset (T1$\rightarrow$PD and PD$\rightarrow$T2). The results with * indicate $p$<0.05 compared with our method based on Wilcoxon signed-rank test.}
\label{tb_one2one}
\begin{tabular}{ccccc} 
\hline\hline
 & \multicolumn{4}{c}{Results [mean \textbf{PSNR} (std), mean \textbf{SSIM} (std)]} \\ 
\cline{2-5}
 & T1$\rightarrow$T2 & FLAIR$\rightarrow$T1 & T1$\rightarrow$PD & ~PD$\rightarrow$T2 \\ 
\hline
pGAN & \begin{tabular}[c]{@{}c@{}}27.10 (1.28)*, \\0.933 (0.019)*\end{tabular} & \begin{tabular}[c]{@{}c@{}}26.91 (1.28)*, \\0.932 (0.016)*\end{tabular} & \begin{tabular}[c]{@{}c@{}}26.42 (0.44)*, \\0.942 (0.009)*\end{tabular} & \begin{tabular}[c]{@{}c@{}}30.30 (0.63)*, \\0.935 (0.034)*\end{tabular} \\
Pix2Pix & \begin{tabular}[c]{@{}c@{}}26.94 (1.28)*, \\0.931 (0.019)*\end{tabular} & \begin{tabular}[c]{@{}c@{}}26.86 (1.33)*, \\0.931 (0.017)*\end{tabular} & \begin{tabular}[c]{@{}c@{}}26.01 (0.47)*, \\0.938 (0.010)*\end{tabular} & \begin{tabular}[c]{@{}c@{}}29.94 (0.45)*, \\\textbf{0.943} (0.019)~\end{tabular} \\
\begin{tabular}[c]{@{}c@{}}Ours \\(SISO)\end{tabular} & \begin{tabular}[c]{@{}c@{}}27.48 (1.33)*, \\0.938 (0.019)*\end{tabular} & \begin{tabular}[c]{@{}c@{}}27.00 (1.24)*, \\0.933 (0.015)*\end{tabular} & \begin{tabular}[c]{@{}c@{}}26.70 (0.42)*, \\0.948 (0.008)~\end{tabular} & \begin{tabular}[c]{@{}c@{}}30.46 (0.47)*, \\0.943\textbf{ }(0.026)~\end{tabular} \\
Ours & \begin{tabular}[c]{@{}c@{}}\textbf{27.78} (1.16), \\\textbf{0.944} (0.017)\end{tabular} & \begin{tabular}[c]{@{}c@{}}\textbf{27.38 }(1.29), \\\textbf{0.938} (0.013)\end{tabular} & \begin{tabular}[c]{@{}c@{}}\textbf{26.96 }(0.48), \\\textbf{0.950} (0.008)\end{tabular} & \begin{tabular}[c]{@{}c@{}}\textbf{30.93}\textbf{ }(0.48), \\0.943\textbf{ }(0.027)\end{tabular} \\
\hline\hline
\end{tabular}
\end{table}

\begin{table}
\renewcommand\tabcolsep{2pt}
\renewcommand\arraystretch{1.0}
\centering
\caption{Quantitative comparison with many-to-one synthesis methods in various tasks on the BraTS dataset (T1+FLAIR$\rightarrow$T2 and T2+FLAIR$\rightarrow$T1) and IXI dataset (T1+T2$\rightarrow$PD and T2+PD$\rightarrow$T1). The results with * indicate $p$<0.05 compared with our method based on Wilcoxon signed-rank test.}
\label{tb_many2one}
\begin{tabular}{ccccc} 
\hline\hline
 & \multicolumn{4}{c}{Results [mean \textbf{PSNR} (std), mean \textbf{SSIM} (std)]} \\ 
\cline{2-5}
 & T1+FLAIR$\rightarrow$T2 & T2+FLAIR$\rightarrow$T1 & T1+T2$\rightarrow$PD & T2+PD$\rightarrow$T1 \\ 
\hline
Hi-Net & \begin{tabular}[c]{@{}c@{}}27.24 (1.33)*, \\0.929 (0.020)*\end{tabular} & \begin{tabular}[c]{@{}c@{}}27.25 (1.62)*, \\0.925 (0.030)*\end{tabular} & \begin{tabular}[c]{@{}c@{}}30.08 (1.54)*, \\0.967 (0.018)*\end{tabular} & \begin{tabular}[c]{@{}c@{}}27.85 (1.12)*, \\0.954 (0.021)*\end{tabular} \\
\begin{tabular}[c]{@{}c@{}}Ours \\(2I3O)\end{tabular} & \begin{tabular}[c]{@{}c@{}}28.41 (1.40)*, \\0.950 (0.016)*\end{tabular} & \begin{tabular}[c]{@{}c@{}}29.17 (1.44)*, \\0.960 (0.013)*\end{tabular} & \begin{tabular}[c]{@{}c@{}}31.76 (0.46)*, \\0.979 (0.005)\end{tabular} & \begin{tabular}[c]{@{}c@{}}29.26 (0.57)*, \\0.968 (0.007)*\end{tabular} \\
Ours & \begin{tabular}[c]{@{}c@{}}\textbf{28.68} (1.29), \\\textbf{0.954} (0.014)\end{tabular} & \begin{tabular}[c]{@{}c@{}}\textbf{29.47} (1.28), \\\textbf{0.963} (0.013)\end{tabular} & \begin{tabular}[c]{@{}c@{}}\textbf{31.86} (0.41), \\\textbf{0.980} (0.004)\end{tabular} & \begin{tabular}[c]{@{}c@{}}\textbf{29.50} (0.51), \\\textbf{0.971} (0.006)\end{tabular} \\
\hline\hline
\end{tabular}
\end{table}

\begin{figure*}[t]
\centerline{\includegraphics[width=0.8\textwidth]{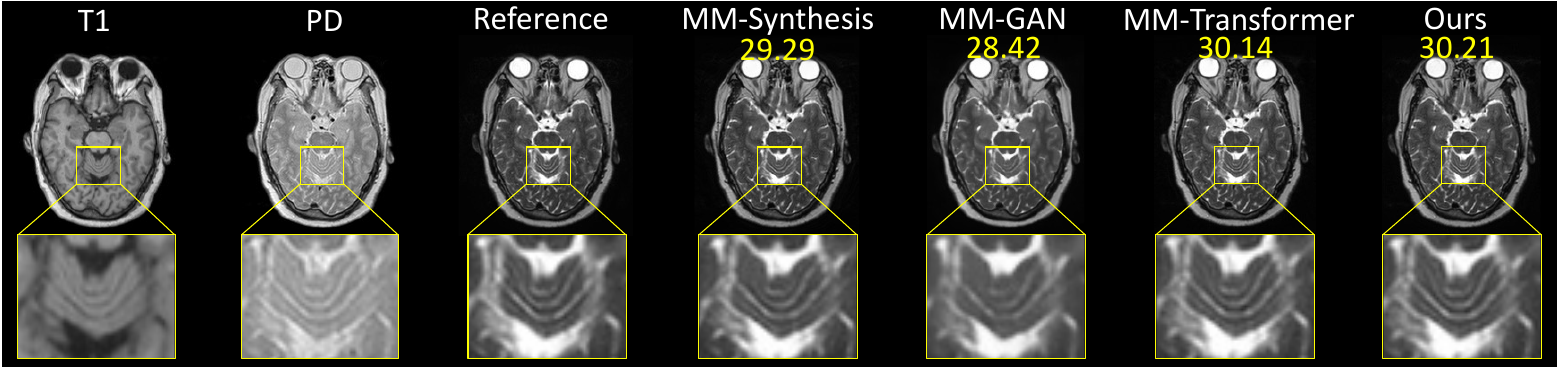}}
\caption{Representative examples (T1+PD$\rightarrow$T2) of unified synthesis on the IXI dataset. Yellow boxes emphasize distinctions between images, which are enlarged for a better view, and the yellow decimals represent PSNR values.}
\label{fig_res_uni_ixi}
%\vspace{-3 mm}
\end{figure*}
% %\vspace{-0.5cm}

\subsubsection{One-to-One Synthesis}
\label{sec_one2one}
For the one-to-one synthesis scenario, two representative synthesis tasks in the BraTS dataset are considered: T1$\rightarrow$T2 and FLAIR$\rightarrow$T1, and another two tasks in the IXI dataset are considered:
T1$\rightarrow$PD and PD$\rightarrow$T2. Note that our unified model can realize one-to-one synthesis during the testing phase (with three missing input modalities zero-padded), however, its training setting is different from the comparison methods that learn mappings between only two modalities. Specifically, our model leverages images from over two modalities (four modalities on the BraTS dataset, and three modalities on the IXI dataset) during the training stage while comparison methods only use images from two modalities for specific tasks. Therefore, to assess the performance of our method under the same training setup, we implement a single-input single-output (SISO) variant of our method that employs two modalities when training. The SISO network contains a single encoding stream, a common encoding stream, five DFUMs, and a single decoding stream.

\begin{table*}
\renewcommand\tabcolsep{3.8pt}
\renewcommand\arraystretch{1.0}
\centering
\caption{Quantitative comparison results of our method and other unified synthesis methods on the BraTS dataset. The results with * indicate $p$<0.05 compared with our method based on Wilcoxon signed-rank test.}
\label{tb_unified_brats}
\begin{tabular}{cccccccc} 
\hline\hline
\multicolumn{4}{c}{Available modalities}                                                                      & \multicolumn{4}{c}{Results [mean \textbf{PSNR} (std), mean \textbf{SSIM} (std)]}                                                                                \\ 
\hline
T1                        & T2                        & T1Gd                      & FLAIR                     & MM-Synthesis                  & MM-GAN                        & MM-Transformer                & Ours                                           \\ 
\hline
                          &                           &                           & \checkmark & 25.30 (1.13)*, 0.895 (0.026)* & 25.17 (1.15)*, 0.896 (0.028)* \  & 25.99 (1.15)*, 0.911 (0.024)* & \textbf{26.18} (1.14)\textbf{, 0.915 }(0.023)  \\
                          &                           & \checkmark &                           & 27.22 (2.11)*, 0.917 (0.040)* & 27.04 (2.40)*, 0.914 (0.046)* & 27.95 (2.22)*, 0.928 (0.037)* & \textbf{28.09 }(2.29),\textbf{ 0.931 }(0.036)  \\
                          & \checkmark &                           &                           & 26.25 (1.37)*, 0.907 (0.031)* & 26.01 (1.23)*, 0.908 (0.027)* & 26.83 (1.09)*,~0.920 (0.023)* & \textbf{27.22 }(1.23),\textbf{ 0.925 }(0.024)  \\
\checkmark &                           &                           &                           & 25.96 (0.71)*, 0.901 (0.024)* & 25.79 (0.82)*, 0.900 (0.028)* & 26.63 (0.79)*, 0.914 (0.023)* & \textbf{26.76 }(0.76),\textbf{ 0.919 }(0.022)  \\
                          &                           & \checkmark & \checkmark & 28.61 (1.62)*, 0.946 (0.016)* & 29.02 (1.59)*, 0.954 (0.013)* & 29.98 (1.33)*, 0.962 (0.010)   & \textbf{30.12} (1.34),\textbf{ 0.963 }(0.009)  \\
                          & \checkmark &                           & \checkmark & 26.72 (1.72)*, 0.924 (0.029)* & 26.56 (1.42)*, 0.927 (0.024)* & 27.61 (1.38)*, 0.940 (0.019)* & \textbf{28.01} (1.46), \textbf{0.943 }(0.020)  \\
\checkmark &                           &                           & \checkmark & 26.34 (0.64)*, 0.917 (0.013)* & 26.86 (0.69)*, 0.928 (0.014)* & 27.80 (0.80)\ , 0.940 (0.012)\    & \textbf{27.88} (0.80),\textbf{ 0.941 }(0.013)  \\
                          & \checkmark & \checkmark &                           & 28.20 (2.35)*, 0.924 (0.042)* & 28.68 (2.31)*, 0.934 (0.036)* & 29.48 (2.34)\ , 0.942 (0.032)   & \textbf{29.51 }(2.35),\textbf{ 0.942 }(0.031)  \\
\checkmark &                           & \checkmark &                           & 26.30 (0.94)*, 0.904 (0.030)* & 26.25 (1.17)*, 0.907 (0.034)* & 27.23 (1.09)*, 0.921 (0.028)* & \textbf{27.35 }(1.13),\textbf{ 0.925 }(0.027)  \\
\checkmark & \checkmark &                           &                           & 25.91 (0.02)*, 0.896 (0.014)* & 26.58 (0.05)*, 0.912 (0.013)* & 27.42 (0.27)*, 0.924 (0.014)*  & \textbf{27.50 }(0.18),\textbf{ 0.926 }(0.012)  \\
                          & \checkmark & \checkmark & \checkmark & 30.60 (1.55)*, 0.967 (0.016)* & 31.06 (1.65)*, 0.970 (0.015)* & 31.92 (1.61)\ , \textbf{0.975} (0.011)   & \textbf{31.94 }(1.94), 0.975 (0.017)  \\
\checkmark &                           & \checkmark & \checkmark & 27.36 (1.27)*, 0.936 (0.019)* & 28.12 (1.12)*, 0.950 (0.013)* & 29.18 (1.33)*, 0.958 (0.013)  & \textbf{29.28 }(1.28),\textbf{ 0.959 }(0.013)  \\
\checkmark & \checkmark &                           & \checkmark & 25.94 (1.16)*, 0.910 (0.023)* & 26.88 (1.32)*, 0.928 (0.024)* & 27.88 (1.56)\ , \textbf{0.941} (0.023)\    & \textbf{27.93 }(1.51), 0.940 (0.025)   \\
\checkmark & \checkmark & \checkmark &                           & 25.87 (1.22)*, 0.883 (0.026)* & 26.64 (1.45)*, 0.902 (0.025)* & 27.30 (1.44)*, 0.913 (0.023)*  & \textbf{27.39 }(1.59),\textbf{ 0.915 }(0.026)  \\
\hline\hline
\end{tabular}
\end{table*}

\begin{table*}
\renewcommand\arraystretch{1.0}
\centering
\caption{Quantitative comparison results of our method and other unified synthesis methods on the IXI dataset. The results with * indicate $p$<0.05 compared with our method based on Wilcoxon signed-rank test.}
\label{tb_unified_ixi}
\begin{tabular}{ccccccc} 
\hline\hline
\multicolumn{3}{c}{Available modalities} & \multicolumn{4}{c}{Results [mean \textbf{PSNR} (std), mean \textbf{SSIM} (std)]} \\ 
\hline
T1 & T2 & PD & MM-Synthesis & MM-GAN & MM-Transformer & Ours \\ 
\hline
 &  & \checkmark & 28.99 (0.33)*, 0.946 (0.016)* & 28.86 (0.85)*, 0.948 (0.008)* & 29.91 (0.72)*, \textbf{0.957} (0.010)\  & \textbf{30.12} (0.81),~0.956 (0.013) \\
 & \checkmark &  & 28.56 (0.74)*, 0.962 (0.006)* & 28.48 (1.35)*, 0.958 (0.011)* & 29.41 (1.17)*, 0.968 (0.007)* & \textbf{29.78 }(1.30),~\textbf{0.970} (0.007) \\
\checkmark &  &  & 25.58 (0.39)*, 0.913 (0.024)* & 25.77 (0.38)*, 0.912 (0.025)* & 26.36 (0.19)*, 0.926 (0.017)* & \textbf{26.70} (0.27),~\textbf{0.930} (0.019) \\
 & \checkmark & \checkmark & 28.74 (0.71)*, 0.964 (0.008)* & 28.32 (0.67)*, 0.961 (0.007)* & 29.41 (0.58)*, 0.970 (0.005)\  & \textbf{29.50} (0.51),~\textbf{0.971} (0.006) \\
\checkmark &  & \checkmark & 29.51 (0.58)*, 0.932 (0.030)* & 30.01 (0.56)*, 0.943 (0.016)* & 31.07 (0.51)*, \textbf{0.952} (0.017)\  & \textbf{31.32} (0.49),~0.948 (0.025) \\
\checkmark & \checkmark &  & 29.55 (0.52)*, 0.969 (0.005)* & 30.56 (0.47)*, 0.973 (0.005)* & 31.45 (0.52)*, 0.978 (0.004)* & \textbf{31.86} (0.41),~\textbf{0.980} (0.004) \\
\hline\hline
\end{tabular}
\end{table*}

Fig.~\ref{fig_res_one2one} gives visual examples of the comparison results. It is shown that our SISO version outperforms PGAN and PixPix slightly, which owes to the network structure including more convolutional channels and attention mechanisms. In one-to-one synthesis scenarios, the DFUM equals a self-attention module, allowing the network to focus on critical regions (\emph{e.g.,} tumor regions) and suppress redundant information. By contrast, our unified method generates images with the most accurate tumor morphology (see Fig.~\ref{fig_res_one2one} (a) and (b)) and subtle structural information (see Fig.~\ref{fig_res_one2one} (c) and (d)). Instead of learning a single mapping between a fixed single input and output like traditional one-to-one methods, our unified model takes a unified learning framework that learns mappings between input and output modalities of various synthesis tasks, and different tasks facilitate each other during training. Take the T1$\rightarrow$T2 task as an example, the related encoding streams $ES_1$, $ES_C$, and decoding stream $DS_2$ are also refined and updated by other tasks including T1$\rightarrow$T1Gd, T1$\rightarrow$FLAIR, T1Gd$\rightarrow$T2, FLAIR$\rightarrow$T2, etc. This makes the network more robust and avoids trapping into a local optimum. Table~\ref{tb_one2one} gives the quantitative results of the comparison. Our method outperforms comparison methods in PSNR and SSIM significantly in most cases based on the Wilcoxon signed-rank test. Both qualitative and quantitative results suggest the superiority of the proposed method.

\subsubsection{Many-to-One Synthesis}
For the many-to-one synthesis scenario, two representative synthesis tasks in the BraTS dataset are considered: T1+FLAIR$\rightarrow$T2 and T2+FLAIR$\rightarrow$T1. Meanwhile, in the IXI dataset, another two many-to-one tasks are considered:
T1+T2$\rightarrow$PD and T2+PD$\rightarrow$T1. 
Since Hi-Net only employs three modalities during training (two source modalities and one target modality), we also implement a two-input three-output (2I3O) variant of our method, which contains two modality-specific encoding streams, a common encoding stream, five DFUMs, and three decoding streams (one for target image synthesis, and two for source image reconstruction, as adopted in Hi-Net). As shown in Fig.~\ref{fig_res_many2one}, synthetic images produced by Hi-Net are more blurry and lose certain high-frequency details (see Fig.~\ref{fig_res_many2one} (a) and (c)). Besides, in Fig.~\ref{fig_res_many2one} (d), Hi-Net does not combine information from multiple modalities effectively and produces streaking noises (emphasized by the red ellipse in Fig.~\ref{fig_res_many2one}). Compared with Hi-Net, our 2I3O version produces better results. The reason is that our method benefits from the multi-modal feature integration strategy of DFUM, which aggregates features through both hard (discrete) and soft (continuous) combinations. As such, DFUM ensures the effective fusion of features while avoiding the loss of valuable information contained in each modality. By comparison, our unified method generates the most visually pleasing images owing to the advantage of DFUM and the unified learning framework (explained in Section~\ref{sec_one2one}).  
Table~\ref{tb_one2one} presents the quantitative comparison results between our method and Hi-Net in four many-to-one synthesis tasks on the BraTS dataset and IXI dataset. It is shown that our unified method achieves the highest values in PSNR and SSIM with $p<0.05$ in most cases, implying that the proposed method has superior performance.

\begin{figure}[t]
\centerline{\includegraphics[width=0.5\textwidth]{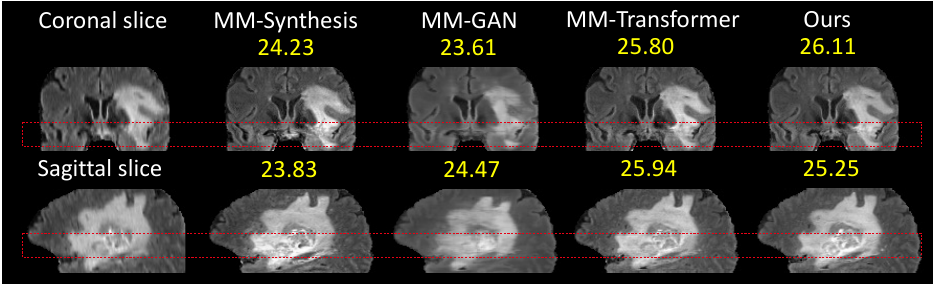}}
\caption{Synthetic FLAIR images (T1+T2+T1Gd$\rightarrow$FLAIR) in coronal and sagittal views. Red dotted boxes emphasize discontinuity in images. The yellow decimals represent PSNR values.}
\label{fig_other_view_brats}
%\vspace{-3 mm}
\end{figure}

\subsubsection{Unified Synthesis}
For the unified synthesis scenario, we compare our method with MM-Synthesis~\cite{chartsias2017multimodal}, MM-GAN~\cite{sharma2019missing}, and MM-Transformer~\cite{liu2023one} with all possible combinations of inputs on two datasets. Concretely, 14 input scenarios are investigated in the BraTS dataset and 6 input scenarios are investigated in the IXI dataset. Table~\ref{tb_unified_brats} and Table~\ref{tb_unified_ixi} summarize the comparison results. The numerical result of each input scenario is averaged on all generated missing modalities. It is demonstrated that the proposed method achieves better values than competing methods in PSNR and SSIM metrics ($p<$0.05) in most input scenarios. 
To show visual examples of the comparison, we select two representative synthesis tasks, including (1) T1+T2+T1Gd$\rightarrow$FLAIR in the BraTS dataset, and (2) T1+PD$\rightarrow$T2 in the IXI dataset, and display the synthetic results in Fig.~\ref{fig_res_uni_brats} and Fig.~\ref{fig_res_uni_ixi}. It is observed that MM-Synthesis may generate excessive contrast for the tumor region (see Fig.~\ref{fig_res_uni_brats}) or lose certain structural information (see Fig.~\ref{fig_res_uni_ixi}). The reason may be that MM-Synthesis is a CNN-based synthesis method that does not use the adversarial loss to constrain the local structure and high-frequency characteristics of the synthetic images. In addition, MM-Synthesis only retains the maximum response value when merging multi-modal features, but the source modality with the highest response value does not necessarily provide sufficient detailed information for the target modality.
MM-GAN tends to produce blurry images that lack rich details (see Fig.~\ref{fig_res_uni_brats} and Fig.~\ref{fig_res_uni_ixi}). The reason may be that in MM-GAN, different input modalities are processed by a single encoding and decoding pathway. This makes the network better at mining modality-invariant features, but less sensitive to modality-specific details, thereby limiting the quality of the generated images. MM-Transformer can effectively employ intra- and inter-contrast information and produce high-quality images. Nevertheless, its synthetic results are slightly inferior to our method. This could be because transformers typically exhibit advantages on large-scale datasets, while our CNN-based network performs better on relatively small datasets (290 cases in the BraTS dataset, and 200 cases in the IXI dataset). In comparison, our method synthesizes images with the highest quality, because we use a more reasonable multi-modal feature integration strategy and employ both modality-shared and modality-specific streams to comprehensively analyze the information from input contrasts.

\subsection{Consistency in Coronal and Sagittal Planes}
\label{sec_consistency}
Since our method primarily focuses on the synthesis of two-dimensional (2D) MR images, we present the synthesis results of axial slices, which are mostly investigated by 2D synthesis works, in previous sections. However, in clinical practice, radiologists also consider information from sagittal and coronal slices for comprehensive disease diagnosis. Considering this, we stack the synthetic axial slices into a 3D volume and observe cross-sections in sagittal and coronal planes. Fig.~\ref{fig_other_view_brats} shows the synthetic results in other views of the same case presented in Fig.~\ref{fig_res_uni_brats}. The intensity consistency of synthetic images from MM-GAN is shown to be the poorest in the other two planes. While synthetic images from the remaining methods may exhibit some intensity discontinuities (due to the inherent limitations of 2D methods), they are visually acceptable.

\begin{table*}
\renewcommand\arraystretch{1.0}
\renewcommand\tabcolsep{8.8pt}
\centering
\caption{Tumor segmentation evaluation on the BraTS dataset. The Dice Coefficient scores (\%) are computed between masks generated by real four-modal sequences and imputed sequences. ``\checkmark" means available real images, and ``$\circ$" means imputed images produced by networks. The results with * indicate $p$<0.05 compared with our method based on Wilcoxon signed-rank test.}
\label{tb_segmentation}
\begin{tabular}{cccclccc} 
\hline\hline
\multicolumn{4}{c}{Input modalities} & \multicolumn{1}{c}{\multirow{2}{*}{Method}} & \multicolumn{3}{c}{Dice Coefficient (\%)} \\ 
\cline{1-4}\cline{6-8}
T1 & T2 & T1Gd & Flair & \multicolumn{1}{c}{} & Whole tumor & Tumor core & Enhancing tumor \\ 
\hline
  & \checkmark & \checkmark & \checkmark & Missing & 90.95 (13.25)* & 82.00 (25.03)* & 87.77 (20.74)* \\
$\circ$ & \checkmark & \checkmark & \checkmark & MM-Synthesis & 97.64 (2.66)* & 92.30 (12.00)* & 87.55 (29.23)* \\
$\circ$ & \checkmark & \checkmark & \checkmark & MM-GAN & 97.65 (3.15)* & 92.65 (8.65)* & 90.60 (21.31)* \\
$\circ$ & \checkmark & \checkmark & \checkmark & MM-Transformer & 97.72 (3.37)* & \textbf{94.55 (7.17)\ } & \textbf{92.93 (21.36)\ } \\
$\circ$ & \checkmark & \checkmark & \checkmark & Ours & \textbf{97.89 (2.57)\ } & 94.07 (7.89)\  & 92.83 (21.35)\  \\ 
\hline
\checkmark &   & \checkmark & \checkmark & Missing & 89.09 (20.83)* & 81.95 (27.09)* & 85.46 (22.01)* \\
\checkmark & $\circ$ & \checkmark & \checkmark & MM-Synthesis & 95.06 (4.96)* & 88.30 (18.75)* & 89.19 (24.00)* \\
\checkmark & $\circ$ & \checkmark & \checkmark & MM-GAN & 96.33 (5.04)* & 91.22 (14.66)* & 90.12 (21.41)* \\
\checkmark & $\circ$ & \checkmark & \checkmark & MM-Transformer & 96.67 (3.98)\  & 92.07 (12.78)* & \textbf{93.21 (21.43)\ } \\
\checkmark & $\circ$ & \checkmark & \checkmark & Ours & \textbf{96.74 (4.02)\ } & \textbf{92.53 (11.55)\ } & 91.35 (22.08)\  \\ 
\hline
\checkmark & \checkmark &   & \checkmark & Missing & 90.60 (9.10)* & 64.88 (25.49)* & 45.66 (28.07)* \\
\checkmark & \checkmark & $\circ$ & \checkmark & MM-Synthesis & 94.39 (4.75)* & 65.29 (20.25)* & 43.55 (26.27)* \\
\checkmark & \checkmark & $\circ$ & \checkmark & MM-GAN & 94.18 (3.95)* & 54.18 (23.73)* & 39.95 (11.95)* \\
\checkmark & \checkmark & $\circ$ & \checkmark & MM-Transformer & \textbf{94.98 (3.79)\ } & 67.92 (19.43)* & 45.87 (23.71)* \\
\checkmark & \checkmark & $\circ$ & \checkmark & Ours & 94.77 (4.51)\  & \textbf{72.55 (18.05)\ } & \textbf{49.20 (26.05)\ } \\ 
\hline
\checkmark & \checkmark & \checkmark &   & Missing & 84.66 (19.94)* & 78.38 (30.70)* & 81.64 (27.79)* \\
\checkmark & \checkmark & \checkmark & $\circ$ & MM-Synthesis & 86.83 (18.87)* & 82.38 (27.18)* & 85.71 (28.74)* \\
\checkmark & \checkmark & \checkmark & $\circ$ & MM-GAN & 82.80 (21.30)* & 80.15 (29.26)* & 85.21 (21.20)* \\
\checkmark & \checkmark & \checkmark & $\circ$ & MM-Transformer & \textbf{89.35 (11.63)\ } & 84.11 (26.38)* & 85.90 (28.77)\  \\
\checkmark & \checkmark & \checkmark & $\circ$ & Ours & 87.46 (19.05)\  & \textbf{84.63 (25.55)\ } & \textbf{85.91 (28.81)\ } \\
\hline\hline
\end{tabular}
\end{table*}

\begin{figure}[t]
\centerline{\includegraphics[width=0.5\textwidth]{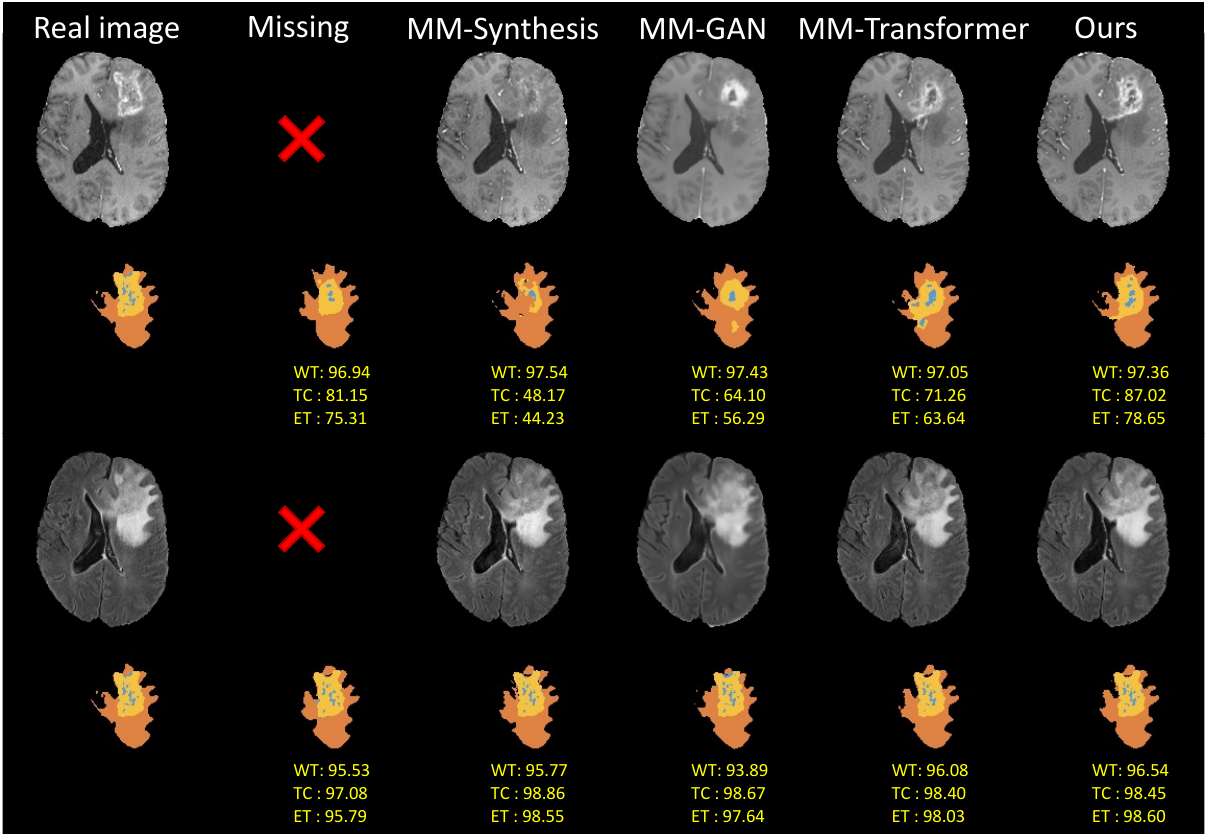}}
\caption{Tumor segmentation with real and imputed T1Gd/FLAIR images. The necrotic and non-enhancing tumor core (NCR/NET) regions are filled in \textcolor{cyan}{blue}, enhancing tumor (ET) regions are filled in \textcolor[rgb]{0.929,0.768,0.356}{yellow}, and peritumoral edema (ED) regions are filled in \textcolor[rgb]{0.823,0.529,0.317}{orange}. Note that the tumor core (TC) consists of NCR/NET and ET; the whole tumor (WT) consists of TC and ED. The yellow decimals represent the Dice Coefficient scores.}
\label{fig_segmentation}
%\vspace{-3 mm}
\end{figure}

\subsection{Tumor Segmentation Evaluation}
In previous sections, we comprehensively presented the image-level quality of the synthetic results. Here, we discuss how the synthetic images further benefit clinical purposes through downstream task evaluation.

Specifically, we choose tumor segmentation as the downstream validation task as it is a crucial step of disease diagnosis and can directly reflect the synthesis accuracy of the tumor regions. Following Liu's work~\cite{liu2023one}, we consider the scenario where a single contrast is missing on the BraTS dataset. First, we use different synthesis methods to obtain imputed four-modal sequences from three available contrasts. 
Then, a U-Net that accepts four-modal inputs is trained for brain tumor segmentation. We report the Dice Coefficient scores of the whole tumor (WT), tumor core (TC), and enhancing tumor (ET) between masks generated by real four-modal sequences and imputed sequences. Besides, to demonstrate the impact of imputed images, we also trained U-Nets that accept three-modal inputs to segment tumors for comparison. The segmentation results are presented in Table~\ref{tb_segmentation}, in which method ``Missing" means using three-modal sequences for segmentation, while other methods provide imputed four-modal sequences for segmentation. It can be observed that in most cases, our method yields the largest performance gains in the accuracy of brain tumor segmentation. This indicates that our method generates the most clinically reliable images and can provide certain assistance for disease diagnosis. Meanwhile, tumor segmentation evaluation reflects that synthetic T1Gd images are crucial for tumor core and enhancing tumor segmentation, while FLAIR images are critical for whole tumor segmentation (more precisely, the peritumoral edema regions within the whole tumor). Fig.~\ref{fig_segmentation} displays visual examples of tumor segmentation with real and imputed T1Gd/FLAIR images. As synthesizing tumor regions in T1Gd images is obviously more challenging, the imputed images can adversely affect tumor core segmentation if the synthesized tumor enhancement pattern does not conform to reality (as seen in the first row of Fig.~\ref{fig_segmentation}).

\subsection{Ablation Study}
In this section, we validate the effectiveness of each newly proposed component in our method.

\subsubsection{The Rationality of the CDS-Encoder}
First, we examine the structural rationality of the CDS-Encoder proposed in our work. To highlight the benefits of our encoding method, we compare it with two other encoding approaches: (1) MMS-Encoder, which employs multiple modality-specific encoding streams, with each stream corresponding to an input modality, and (2) C-Encoder, which uses a single common encoding stream to handle all possible input modalities. Table~\ref{tb_ablation} displays the quantitative comparison results, where all variants have the same network structure except for the encoder. We report the average numerical results of each method across 14 input scenarios on the BraTS dataset. It is shown that the CDS-Encoder achieves 28.23 dB in PSNR and 0.937 in SSIM, outperforming the MMS-Encoder by +0.40 dB in PSNR and +0.006 in SSIM, and C-Encoder by +0.98 dB in PSNR and +0.012 in SSIM. These superior results demonstrate that our proposed CDS-Encoder is both rational and effective for processing multi-modal inputs.

\begin{table}
\renewcommand\arraystretch{1.0}
\renewcommand\tabcolsep{5.8pt}
\centering
\caption{Ablation study on the BraTS dataset. All results are averaged across 14 input scenarios. The results with * indicate $p$<0.05 compared with our method based on Wilcoxon signed-rank test.}
\label{tb_ablation}
\begin{tabular}{lcc} 
\hline\hline
Experiments & \textbf{PSNR} & \textbf{SSIM} \\ 
\hline
MMS-Encoder & 27.83 (1.45)* & 0.931 (0.019)* \\
C-Encoder & 27.25 (0.57)* & 0.925 (0.010)* \\
CDS-Encoder (ours) & \textbf{28.23} (1.46)\  & \textbf{0.937} (0.018)\  \\ 
\hline
Max operation & 27.79 (1.10)* & 0.932 (0.015)* \\
HeMIS & 27.60 (0.93)* & 0.930 (0.013)* \\
TFusion & 27.87 (1.48)* & 0.932 (0.019)* \\
DFUM (ours) & \textbf{28.23} (1.46)\  & \textbf{0.937} (0.018)\  \\ 
\hline
Ours w/o CL & 27.93 (1.48)* & 0.933 (0.018)* \\
Ours w/ CL & \textbf{28.23} (1.46)\  & \textbf{0.937} (0.018)\  \\ 
\hline
Baseline & 27.17 (1.43)* & 0.923 (0.021)* \\
Baseline+CDS-Encoder & 27.51 (1.53)* & 0.928 (0.022)* \\
Baseline+CDS-Encoder+DFUM & 27.93 (1.48)* & 0.933 (0.018)* \\
Baseline+CDS-Encoder+DFUM+CL & \textbf{28.23} (1.46)\  & \textbf{0.937} (0.018)\  \\
\hline\hline
\end{tabular}
\end{table}

\subsubsection{The Effectiveness of DFUM}
So far, there are many methods available for fusing multi-modal features, including arithmetic strategies (through arithmetic
function such as averaging~\cite{dorent2019hetero, lau2019unified}), selection strategies (through Max, Min operator~\cite{chartsias2017multimodal})
and attention-based strategies~\cite{chen2019robust,zhou2021latent,liu2022tfusion} (through weighted sum), etc. In this work, we propose a feature unification strategy (the DFUM) to prepare unified latent features for the decoding stage, which consists of soft feature integration and hard feature integration. To demonstrate the superiority of DFUM, we compare it with three feature unification strategies: (1) Max operation~\cite{chartsias2017multimodal}, which only retains the maximum value of input multi-modal features at spatial positions, (2) HeMIS~\cite{havaei2016hemis}, which computes the mean and variance of input multi-modal features, and (3) TFusion~\cite{liu2022tfusion}, a transformer-based N-to-one fusion method that generates weight maps for each contrast through transformer layers. Table~\ref{tb_ablation} shows the quantitative comparison results on the BraTS dataset. 
From the table, the proposed DFUM outperforms Max operation by +0.44 dB in PSNR and +0.005 in SSIM, HeMIS by +0.63 dB in PSNR and +0.007 in SSIM, and TFusion by +0.36 dB in PSNR and +0.005 in SSIM. Although TFusion adopts an advanced transformer layer for feature fusion, it needs to project the input feature into a compressed vector before applying multi-head self-attention, which loses information and limits performance. Compared with the three methods, our DFUM achieves the best results, indicating that DFUM can most effectively combine features from available modalities.

\subsubsection{The Effectiveness of Curriculum Learning}
As mentioned in Section~\ref{training_scheme}, we utilize a curriculum learning (CL) strategy during training to allow the network to learn synthetic tasks from easy to difficult. Here, we verify the effectiveness of this training scheme
by conducting two experiments: ``Ours w/ CL" which incorporates the CL scheme in the training, and ``Ours w/o CL" which does not use the CL strategy, and exposes the network to random Availability Conditions at all times. The results in Table~\ref{tb_ablation} indicate that incorporating CL increases the performance by +0.30 dB in PSNR and +0.004 in SSIM, demonstrating the effectiveness of the CL training scheme.

\subsubsection{Step-wise Ablation}
We also conduct a step-wise ablation study by progressively adding the network components in order. For this purpose, we start from a baseline network that adopts multiple encoding streams for encoding (MMS-Encoder), Max operation for feature fusion, and exposure to random input-output configurations for training. Then, we gradually involve the encoder into the CDS-Encoder, involve the Max fusion into DFUM, and finally add the CL training strategy. The experimental results are shown in Table~\ref{tb_ablation}. Performance gains are observed as each component is added sequentially, which implies the relative importance of the CDS-Encoder, DFUM, and CL scheme.

\section{Discussion}
In this paper, we propose a unified framework for multi-modal MR image synthesis, which can take arbitrary available contrasts as input and impute the missing contrasts. Section~\ref{sec_results} demonstrates the effectiveness of each module of our method and its superiority over state-of-the-art methods. In this section, we further discuss its potential in 3D volume synthesis, limitations, and future works.

\begin{figure}[t]
\centerline{\includegraphics[width=0.5\textwidth]{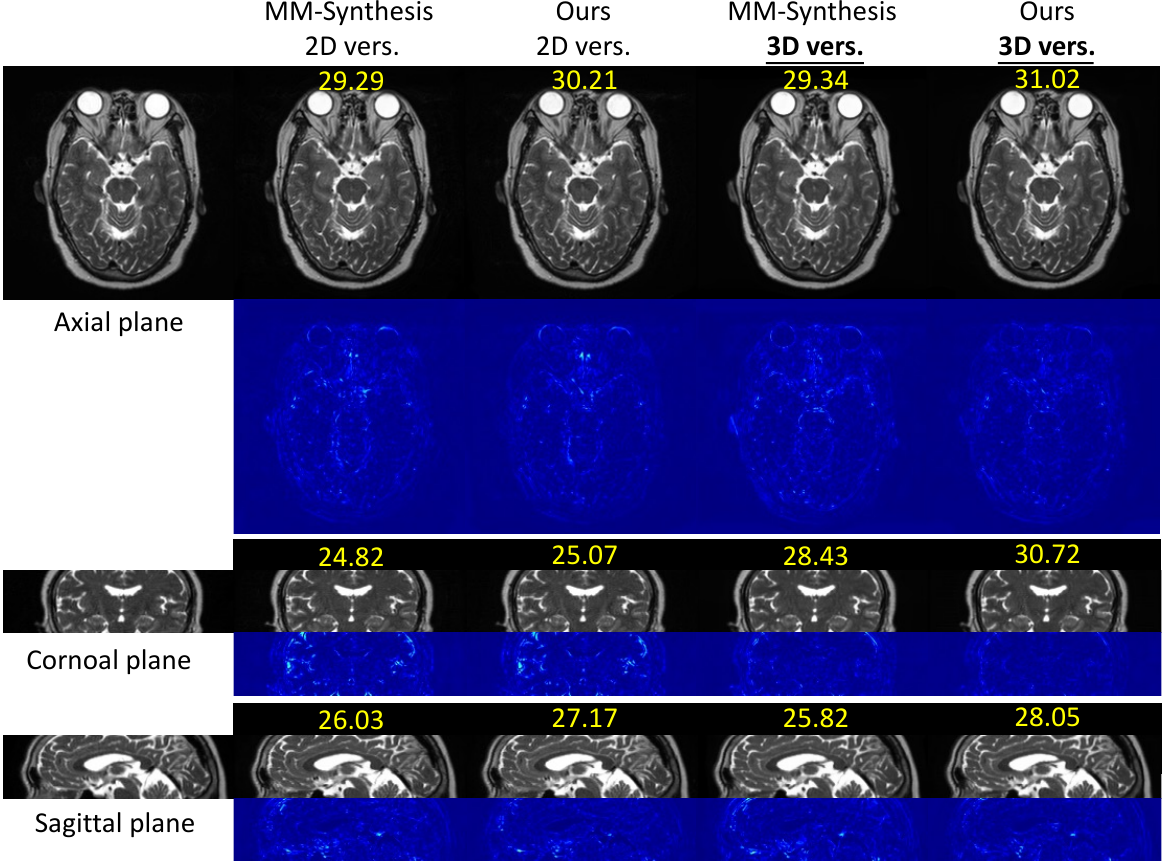}}
\caption{Synthetic T2 images (T1+PD$\rightarrow$T2) in axial, coronal and sagittal views. The second, fourth, and sixth rows present difference maps between synthetic images and real images, which help recognize subtle differences. The yellow decimals represent PSNR values.}
\label{fig_3d_compare}
%\vspace{-3 mm}
\end{figure}

\begin{table}[t]
\renewcommand\tabcolsep{5.8pt}
\renewcommand\arraystretch{1.0}
\centering
\caption{Quantitative comparison results of our method (3D vers.) and MM-Synthesis (3D vers.) on the IXI dataset regarding task T1+PD$\rightarrow$T2. The results with * indicate $p$<0.05 compared with our method based on Wilcoxon signed-rank test.}
\label{tb_3d}
\begin{tabular}{lcc} 
\hline\hline
Method & PSNR (dB) & SSIM \\ 
\hline
MM-Synthesis (3D vers.) & 29.64 (0.80)* & 0.917 (0.017)* \\
Ours (3D vers.) & \textbf{31.85 (0.48)\ } & \textbf{0.940 (0.024)\ } \\
\hline\hline
\end{tabular}
\end{table}

\subsection{Potential in 3D Volume Synthesis}

In Section~\ref{sec_consistency}, we discussed the image quality of 2D methods in other planes. Although images obtained by stacking synthetic axial slices show acceptable consistency in intensity values, there still exist flaws due to the lack of 3D spatial information. Therefore, we here explore the potential of extending our method to the 3D synthesis scenario, which can directly synthesize volumetric data.

To this end, we conduct preliminary experiments on the IXI dataset. We first replace 2D convolutional layers with 3D ones in our network. The modified 3D network takes as input three-modal volumetric data of size 3$\times$32$\times$224$\times$224, with each modality being a randomly cropped patch of size 32$\times$224$\times$224 (the missing modalities are zero-imputed). The network output, of size 3$\times$32$\times$224$\times$224, contains the reconstruction of available modalities and the synthetic results of missing modalities. Likewise, a 3D version of MM-Synthesis is implemented for comparison.

During the training phase, the batch size is set to 2. We use sliding window and stitching operations during the testing phase to obtain the final synthetic volume of size 3$\times$60$\times$224$\times$224. Note that the volumetric data only contains the central 60 slices as it was originally prepared for 2D synthesis as stated in Section~\ref{sec_material}. In Fig.~\ref{fig_3d_compare} and Table~\ref{tb_3d}, qualitative and quantitative results are presented regarding a representative synthesis task ``T1+PD$\rightarrow$T2". Observe that our 3D version significantly surpasses MM-Synthesis (3D vers.) in both PSNR and SSIM. Furthermore, compared with the 2D synthesis results in Table~\ref{tb_unified_ixi}, the images synthesized by our 3D method show an obvious improvement in PSNR (from 31.32 dB to 31.85 dB), despite a slight decrease in SSIM (from 0.948 to 0.940).

\subsection{Limitations and Future Works}
Apart from the aforementioned discontinuity issues in coronal and sagittal planes, our method also has the following limitations: First, the use of modality-specific encoding and decoding streams in the network leads to potential memory consumption issues when extending the framework to over four modalities or to 3D scenarios. Specifically, the parameter count for the four-modal model on the BraTS dataset is 90.74M, including 80.20M for the generator and 10.54M for the discriminator; while on the IXI dataset, the parameter count for the three-modal model is 75.06M, including 67.16M for the generator and 7.90M for the discriminator. This indicates that adding a new modality will increase the parameters by 15.68M.
As for 3D synthesis, although our preliminary experiments on 3D synthesis have shown some effectiveness, the preliminary version can only support a small batch size of 2. Second, it is not yet flexible enough to achieve multi-modal image synthesis across different datasets. An ideal unified multi-modal image synthesis method aims to train jointly on multi-site datasets, obtaining a single model that works on different datasets. However, the current method can only train separate models for different datasets. 

According to these limitations, we can pursue the following research directions in the future. First, designing a synthesis network with lightweight architectures. Possible solutions may include model pruning~\cite{liu2018rethinking}, knowledge distillation~\cite{gou2021knowledge}, etc. Second, developing a more universal and flexible cross-dataset model for multi-modal image synthesis. The challenges to be addressed in joint training on multiple datasets include domain shifts among different datasets, and how to integrate synthesis tasks of different modalities across various datasets.

\section{Conclusion}
In this paper, we present a novel approach for unified multi-modal image synthesis using a generative adversarial network. To fully exploit the commonality and discrepancy information of available modalities, we introduce a Commonality- and Discrepancy-Sensitive Encoder for the generator that analyses both modality-invariant and modality-specific information, respectively. Besides, we devise a Dynamic Feature Unification Module that can effectively derive the unified features from a varying number of available modalities. Comprehensive experiments on the BraTS dataset and IXI dataset demonstrate the superiority of our method over state-of-the-art synthesis methods.

\bibliographystyle{IEEEtran}

\bibliography{main}
\end{document}